\documentclass[10pt,journal,compsoc]{IEEEtran}

\usepackage{graphicx}
\usepackage{amsmath}
\usepackage{amssymb,subfigure,cases}
\usepackage{setspace}
\usepackage{color}
\usepackage{hyperref}
\usepackage{algorithm}
\usepackage{algorithmicx}
\usepackage{algpseudocode}
\usepackage{multirow}
\usepackage{booktabs} 
\usepackage{url}
\usepackage{bm}
\usepackage{verbatim}
\usepackage[dvipsnames]{xcolor}

\newcommand{\R}{\mathbb{R}}

\ifCLASSOPTIONcompsoc
\usepackage[nocompress]{cite}
\else
\usepackage{cite}
\fi

\ifCLASSINFOpdf
\else
\fi

\begin{document}
%
\title{Disentangled Latent Transformer for Interpretable Monocular Height Estimation}

\author{
     Zhitong~Xiong, Sining Chen, Yilei Shi, \IEEEmembership{Member,~IEEE}, and Xiao Xiang Zhu, \IEEEmembership{Fellow,~IEEE}
\IEEEcompsocitemizethanks{
\IEEEcompsocthanksitem Z. Xiong, S. Chen and X. X. Zhu are with the Data Science in Earth Observation
(SiPEO, formerly: Signal Processing in Earth Observation), Technical University of Munich (TUM), 80333 Munich, Germany. S. Chen and X. X. Zhu are also with the Remote Sensing Technology Institute (IMF), German Aerospace Center (DLR), 82234 We{\ss}ling, Germany (e-mails: zhitong.xiong@tum.de; sining.chen@dlr.de;  xiaoxiang.zhu@dlr.de)

\IEEEcompsocthanksitem Y. Shi is with the Chair of Remote Sensing Technology, Technical University of Munich (TUM), 80333 Munich, Germany. (e-mail: yilei.shi@tum.de)
}}


\IEEEtitleabstractindextext{
	\begin{abstract}
		Monocular height estimation (MHE) from remote sensing imagery has high potential in generating 3D city models efficiently for a quick response to natural disasters. Most existing works pursue higher performance. However, there is little research exploring the interpretability of MHE networks. In this paper, we target at exploring how deep neural networks predict height from a single monocular image. Towards a comprehensive understanding of MHE networks, we propose to interpret them from multiple levels: 1) Neurons: unit-level dissection. Exploring the semantic and height selectivity of the learned internal deep representations; 2) Instances: object-level interpretation. Studying the effects of different semantic classes, scales and spatial contexts on height estimation; 3) Attribution: pixel-level analysis. Understanding which input pixels are important for the height estimation. 
		Based on the multi-level interpretation, a disentangled latent Transformer network is proposed towards a more compact, reliable and explainable deep model for monocular height estimation. Furthermore, a novel unsupervised semantic segmentation task based on height estimation is first introduced in this work. Additionally, we also construct a new dataset for joint semantic segmentation and height estimation. Our work provides novel insights for both understanding and designing MHE models. The dataset and code are publicly available at \url{https://github.com/ShadowXZT/DLT-Height-Estimation.pytorch}.
		
	\end{abstract}
}

\maketitle

\IEEEpeerreviewmaketitle
\section{Introduction}
\IEEEPARstart{U}{rban} disaster monitoring has a close relationship with the lives of residents living in cities. The geometric information of 3D cities can be useful for urban planning, damage monitoring, disaster forecasting, and so on. In this context, obtaining the geometric information efficiently from remote sensing imagery is essential for a rapid response to time-critical world events, e.g., natural hazards and damage assessment. Estimating the geometry from images are long-standing goals of remote sensing and Earth observation \cite{zhu2017deep}. LiDAR, radar, stereo photogrammetry require expensive devices and complicated data processing, while predicting height from a single image is a low-cost and timely-updated way.
	
\begin{figure*}
	\begin{center}
		\includegraphics[width=0.98\textwidth]{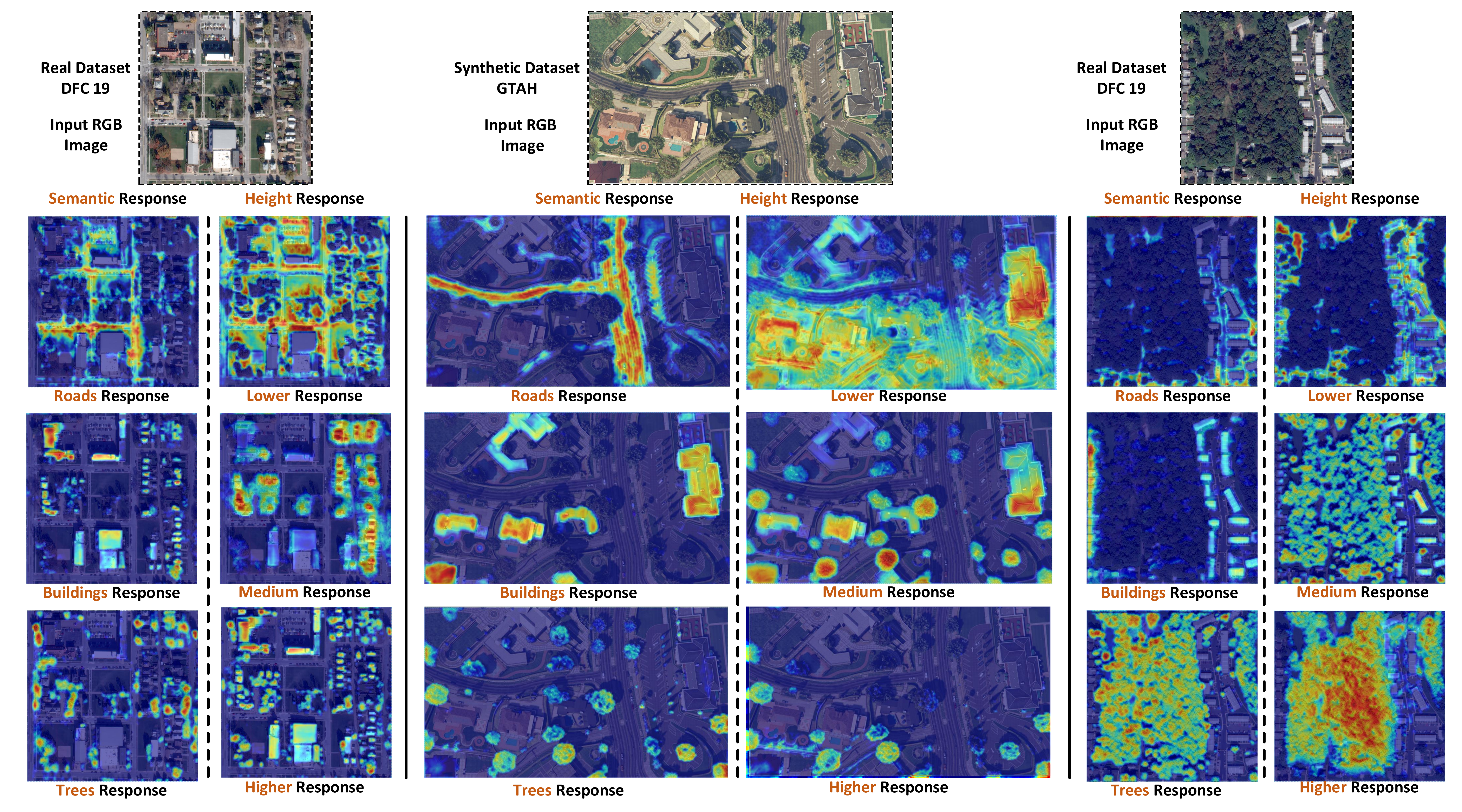}
	\end{center}
	\caption{MHE networks learn to recognize different semantic objects (road, building and tree) and height ranges implicitly. This figure shows the strong selectivity of Transformer-based MHE networks on both the GTAH dataset and the real-world DFC 2019 dataset. (Best viewed with zoom in.)}
	\label{fig::motvation}
\end{figure*}

Motivated by the development of monocular depth estimation (MDE) task \cite{eigen2014depth,fu2018deep,kuznietsov2017semi}, various methods have been proposed for monocular height estimation. Mou et. al \cite{mou2018im2height} designed a residual convolutional network for height estimation and demonstrated its effectiveness on the instance segmentation task. Christie et al. \cite{christie2020learning,christie2021geocentricpose} proposed to estimate the geocentric pose from monocular oblique images. Since estimating height from monocular images is an ill-posed task, improving the interpretability and reliability of deep models is the highest priority for risk-sensitive applications. However, there are few works focusing on understanding deep networks for MHE. 
\begin{figure}
	\centering
	\includegraphics[width=0.5\textwidth]{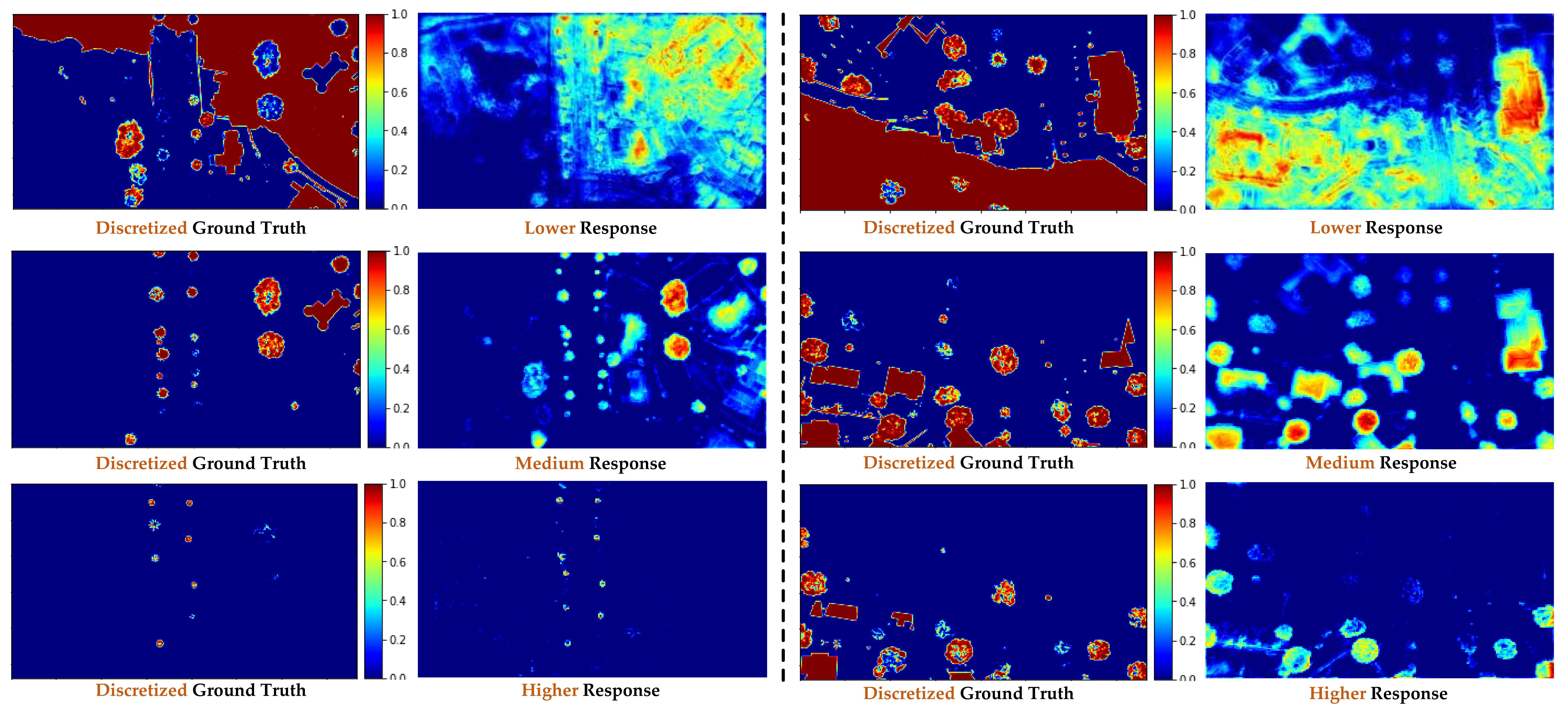}
	\caption{Visualization of the high correlation between height ranges and feature maps of MHE networks. (Best viewed with zoom in)}
	\label{height-selectivity}
\end{figure}

As for interpretable deep models, previous research on image classification \cite{zhang2018interpretable,chen2018looks} and object detection \cite{wu2019towards} has been studied. However, MHE is a dense prediction task involving pixel-wise regression. This makes image-level and object-level interpretation methods not applicable. The most relevant task to MHE is MDE. Dijk et al. \cite{dijk2019neural} investigated important visual cues in input monocular images for depth estimation. Hu et al. \cite{hu2019visualization} tried to find the most relevant sparse pixels for estimating the depth. You et al. \cite{you2021interpretable} first found the depth selectivity of some hidden units, which showed promising insights for interpreting MDE models. However, only focusing on the unit-level or global explanations neglects the semantic content of input images, which is insufficient to interpret complex and out-of-distribution scenarios.

\begin{figure*}
	\centering
	\includegraphics[width=0.98\textwidth]{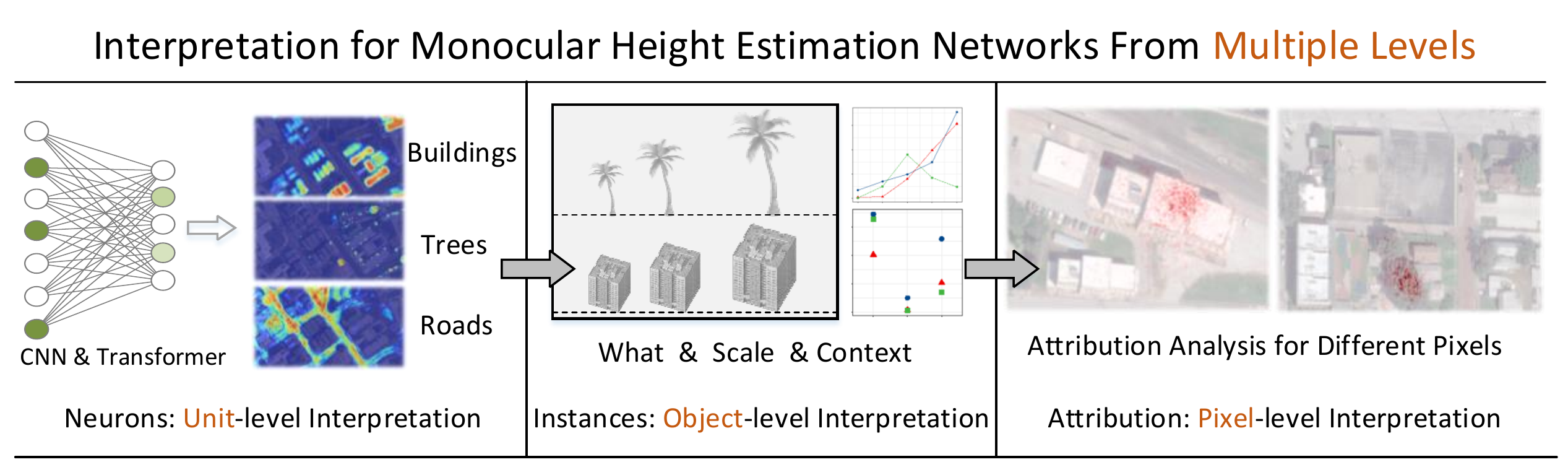}
	\caption{The whole pipeline of the proposed multi-level interpretation framework. Three levels are considered: 1) Neurons: unit-level network dissection; 2) Instances: object-level interpretation; 3) Attribution analysis: pixel-level interpretation.}
	\label{mainarch}
\end{figure*} 

Although MHE shares some similar characteristics with MDE, there are still several aspects making them quite different. Firstly, height is an inherent attribute of objects, which should not change under different views. While the depth of objects highly depend on the camera pose. Secondly, estimating height from top-view may be far more ambiguous than depth estimation from street-view because of the severe occlusion and lack of textures. Thirdly, object types, scales and scene layouts vary greatly in remotely-sensed imagery. These differences also make it not suitable to apply interpretation methods of MDE directly to MHE.

Considering the aforementioned issues, in this work, we aim at 1) understanding deep networks for MHE from multiple levels; 2) designing reliable and explainable deep models for MHE task. Specifically, we propose to explain the deep MHE model from the following three levels:

1) \textbf{Neurons}: unit-level network dissection. Exploring the properties of hidden units to understand what internal representations the MHE model has learned. This is a fundamental component towards an inherent interpretable machine learning model. From this perspective, we find that MHE networks learn disentangled representations to different semantic concepts and height ranges. As shown in Fig. \ref{fig::motvation} and Fig. \ref{height-selectivity}, roads, trees and buildings are automatically recognized, and they are also selective to different height ranges. 

2) \textbf{Instances}: object-level interpretation. Based on the understanding of neurons in MHE networks, we study the effects of changes in semantic objects on MHE networks. By observing the behavior of MHE networks against the changes of these factors, we find that semantic classes, scales and contexts are the main factors influencing the height prediction results.

3) \textbf{Attribution}: pixel-level analysis. Towards a comprehensive understanding of the MHE model, it is necessary to know which input pixels are responsible for the height estimation. Based on the understanding of neurons in MHE networks, we propose to attribute the height prediction locally to its input images. Additionally, both Transformer and CNN are analyzed and compared using local attribution analysis.

Considering that there are limited datasets that contain high resolution height maps and pixel-wise semantic labels, we construct a new dataset, named Washington DC (WDC) dataset, to foster research on improving both the interpretability and performance of MHE models.
To the best of our knowledge, this work is the first attempt to analyze how deep networks work on the monocular height estimation task. Our contributions can be summarized as follows.
\begin{itemize}
\item[(1)] We find that deep neurons learned by MHE models are highly selective to both height ranges and semantic classes. Based on the high class-selectivity of MHE networks, we propose a simple yet effective out-of-distribution detection method for MHE task.

\item[(2)] Based on the observed class-selectivity, we conduct object-level experiments and find that semantic classes, object scales, spatial contexts are main factors influencing height prediction.

\item[(3)] We compare Transformer and CNN-based models on MHE task by the local attribution analysis. Results reveal that Transformer-based networks have stronger height and class-selectivity, and can learn more effective contexts than CNNs.

\item[(4)] Based on the multi-level interpretation, we propose a novel disentangled latent Transformer model for learning disentangled representations towards a more explainable and efficient MHE model. 

\item[(5)] A novel unsupervised semantic segmentation task is first introduced in this work based on the understanding of height estimation networks. Additionally, a new dataset that contains RGB and nDSM (normalized Digital Surface Models) modality is constructed for joint semantic segmentation and height estimation.
\end{itemize}

\section{Related Work}
We review related works in this section from the following three aspects: 
\subsection{Monocular Height Estimation}
With the development of deep learning, various methods have been proposed for MHE. Srivastava et al. \cite{srivastava2017joint} proposed to predict height and semantic labels jointly in a multi-task deep learning framework. Mou et al. \cite{mou2018im2height} designed a residual CNN for height estimation and demonstrated its effectiveness on instance segmentation task. Besides, Conditional generative adversarial network (cGAN) \cite{ghamisi2018img2dsm} was proposed to frame height estimation as an image translation task. Kunwar et al. \cite{kunwar2019u} exploited semantic labels as priors to enhance the performance of height estimation on the large-scale Urban Semantic 3D (US3D) dataset \cite{bosch2019semantic}. Xiong et al. \cite{xiong2021benchmark} designed and constructed a large-scale benchmark dataset for cross-dataset transfer learning on the height estimation task, which includes a large-scale synthetic dataset and several real-world datasets. Swin Transformer \cite{liu2021swin} was used in their work for transferable representation learning on MHE task. Monocular height estimation can be widely used in high-risk Earth observation tasks, while there is no research that focuses on the interpretability of deep learning models for MHE task.

\subsection{Methods for Understanding Monocular Depth Estimation}
Recently, monocular depth estimation has been studied extensively. With the surge of deep learning, extensive methods have been designed to obtain better performance, including geometry-constrained learning \cite{yin2019enforcing}, multi-scale \cite{eigen2014depth}, multi-task learning methods \cite{zhu2020edge}. To understand what these MDE networks have learned, Dijk et al. \cite{dijk2019neural} studied important visual clues used by deep networks when predicting the depth. They mainly focused on object-level interpretation of MDE networks. Hu et al. \cite{hu2019visualization} attempted to find only a selected sparse set of image pixels to estimate depth. A separate network is designed to predict those sparse pixels. However, these methods neglect the inherent representations that the model have learned. You et al. \cite{you2021interpretable} first found the depth selectivity of some hidden units, which showed promising insights for interpreting MDE models. However, merely focusing on hidden units for model interpretation ignores the semantic content of input images.

\subsection{Interpretable and Explainable Deep Neural Networks}
\textbf{Interpretable deep models.} For image classification and object detection tasks, several methods attempted to design inherently interpretable models. Chen et al. \cite{chen2018looks} proposed to find prototypical parts and explain the reason for making final decisions. Towards the interpretability for person re-identification task, Liao et al. \cite{liao2020interpretable} designed a model to make the matching process of feature maps explicit. Zhang et al. \cite{zhang2018interpretable} designed interpretable CNNs by making each filter represent a specific object part. Liang et al. \cite{liang2020training} trained interpretable CNNs by learning class-specific deep filters, namely, encouraging each filter only to account for few classes. Similarly, You et al. \cite{you2021interpretable} proposed to improve the depth selectivity by designing specific loss functions for MDE models.

\textbf{Explainable deep networks.} Many researchers focused on saliency-based and attribution-based methods \cite{sundararajan2017axiomatic} for explainable deep networks. They aimed to highlight which pixels of input images are important for predicting the result \cite{selvaraju2017grad,zeiler2014visualizing,wang2019learning}. 
However, attribution and saliency-based methods are not directly applicable to dense prediction tasks, including MDE and MHE, since it is not reasonable to highlight all the pixels to attribute the dense prediction globally. Gu et al. \cite{gu2021interpreting} proposed a local attribution method for interpreting super-resolution networks. They chose to interpret features instead of pixels for super-resolution task, which inspires our work on pixel-wise attribution for MHE networks.

In this paper, our proposed method considers multi-level factors to give a more comprehensive understanding of the MHE task, which differs from the above approaches.
\begin{figure}[t]
	\centering
	\includegraphics[width=0.485\textwidth]{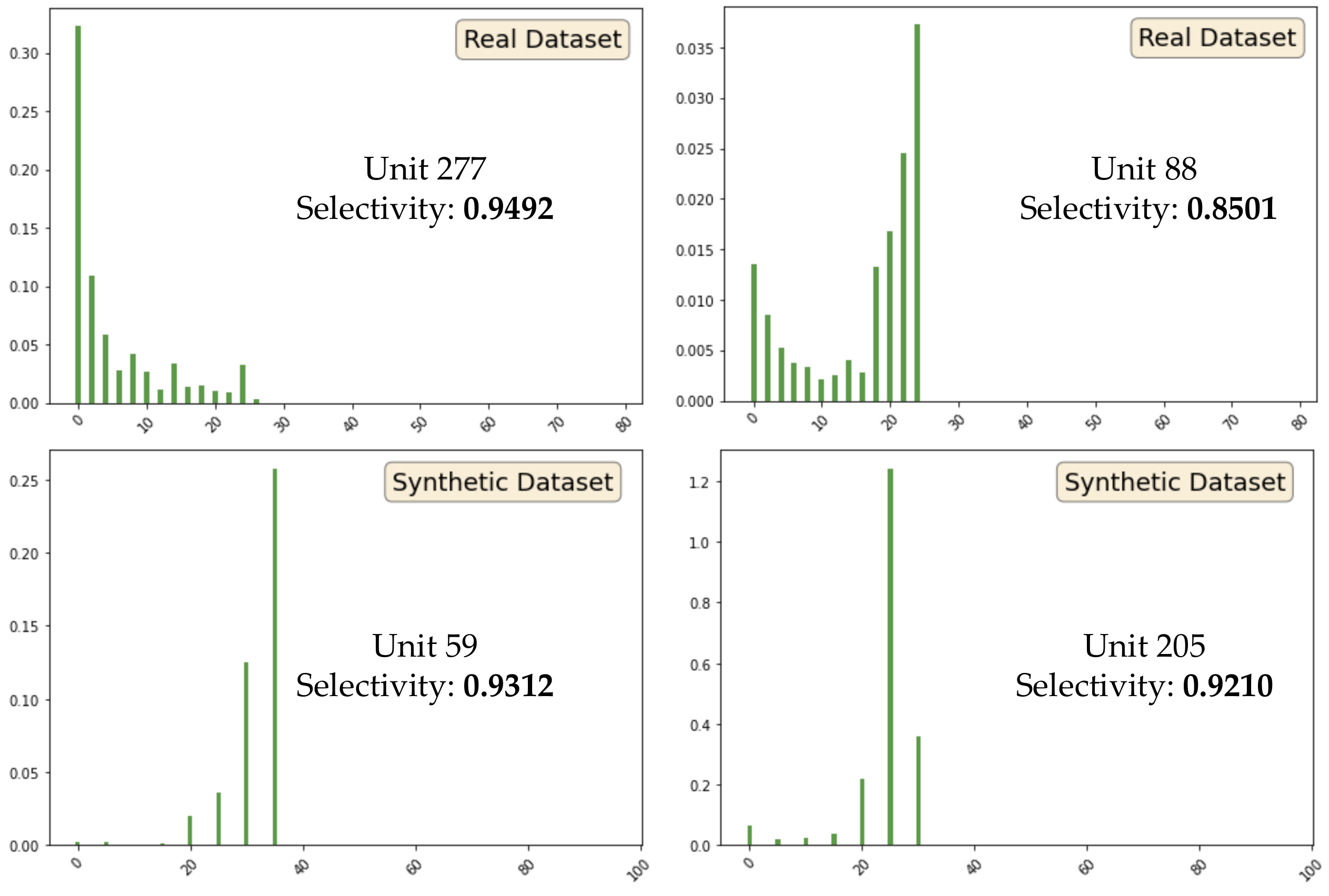}
	\caption{Visualization of the average response $\text{HR}$ for units with large height-selectivity on both the DFC 2019 dataset and the GTAH dataset.}
	\label{HSB}
\end{figure}

\begin{figure}
	\centering
	\includegraphics[width=0.484\textwidth]{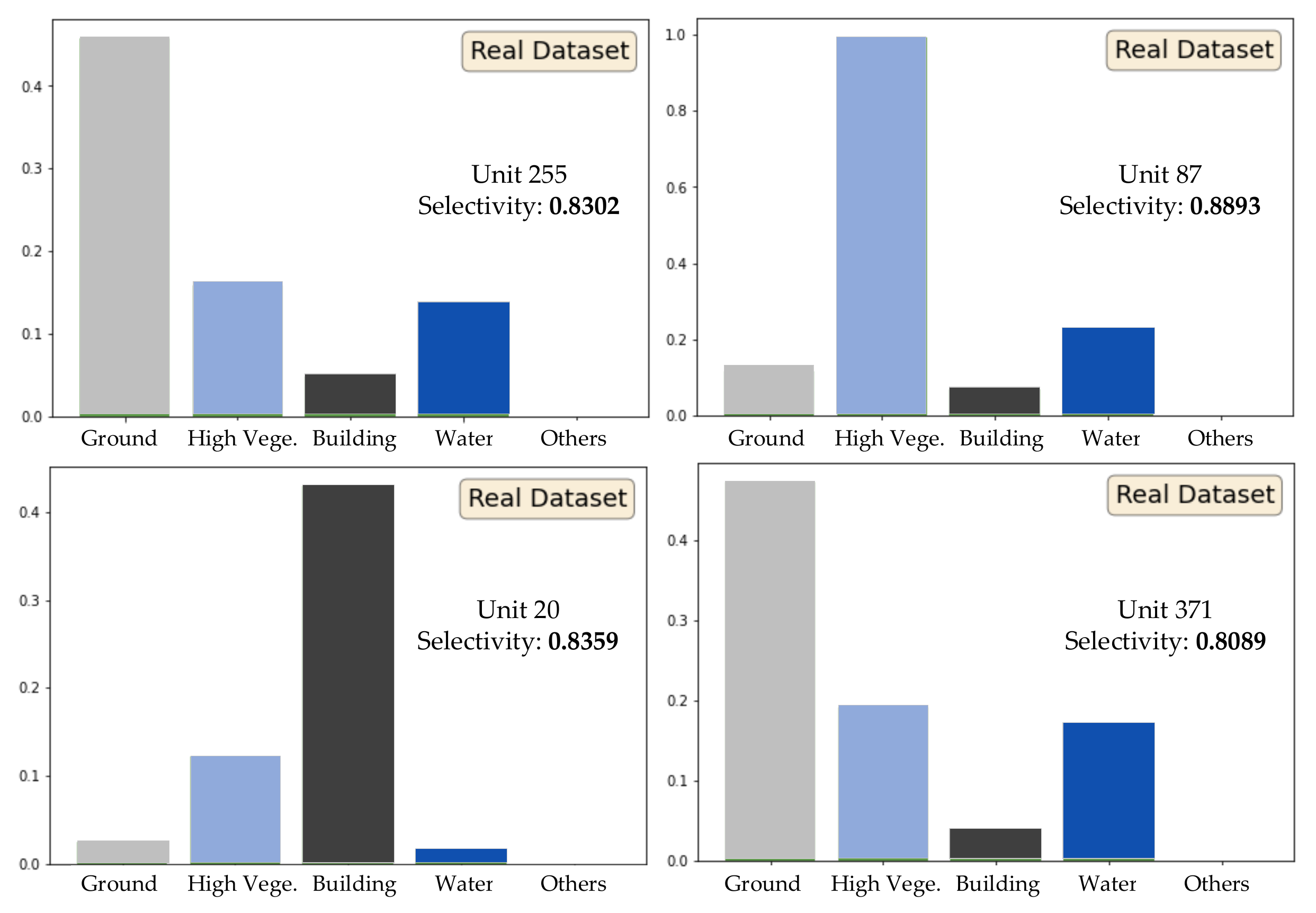}
	\caption{Visualization of the average responses for units with large class-selectivity.}
	\label{CS-Real}
\end{figure} 

\section{Methodology}
The motivation of this work is two-folds: 1) interpreting the behavior of MHE networks from multiple levels for a better understanding of MHE models; 2) designing reliable and explainable deep models for MHE task. The whole pipeline for illustrating the multi-level interpretability framework is presented in Fig. \ref{mainarch}. To understand the behavior of learned hidden units, we examine the learned knowledge of MHE networks by visualizing the representation properties of specialized deep filters. Based on the finding that semantic classes are highly related to height predictions, we then study the effects of different semantic objects and properties on MHE networks. Finally, we design a pixel-level local attribution method for interpreting MHE networks based on the finding in unit-level and object-level interpreting results.

In this section, we will introduce the proposed multi-level interpretation framework as well as the proposed disentangled latent Transformer model in detail.
\subsection{Unit-level Interpretation of MHE Models}
\label{Unit}
Objects of different semantic types usually have distinct height attributes. Thus, the geometric information in height maps should have high correlation with the semantic information \cite{kunwar2019u}. Inspired by this, we choose to examine the learned 
interior activations to find human-understandable representations. From numerous visualizations of deep activations, we find that some units of the model are highly selective to different semantic classes and height ranges. As shown in Fig. \ref{fig::motvation}, different semantic objects including road, building and tree are localized accurately in an implicit manner with only height supervision. This finding supports the assumption that semantic information and geometric information are highly correlated.

Moreover, we also find that some neurons are selective to different height ranges, which shares the same conclusion with \cite{you2021interpretable} for MDE task. As displayed in Fig. \ref{height-selectivity}, the left columns show masks of different discretized depth ranges, and the right columns are feature maps of the MHE network. Obviously, we can see a very high correlation between them.
\begin{figure*}[!]
	\centering
	\includegraphics[width=0.96\textwidth]{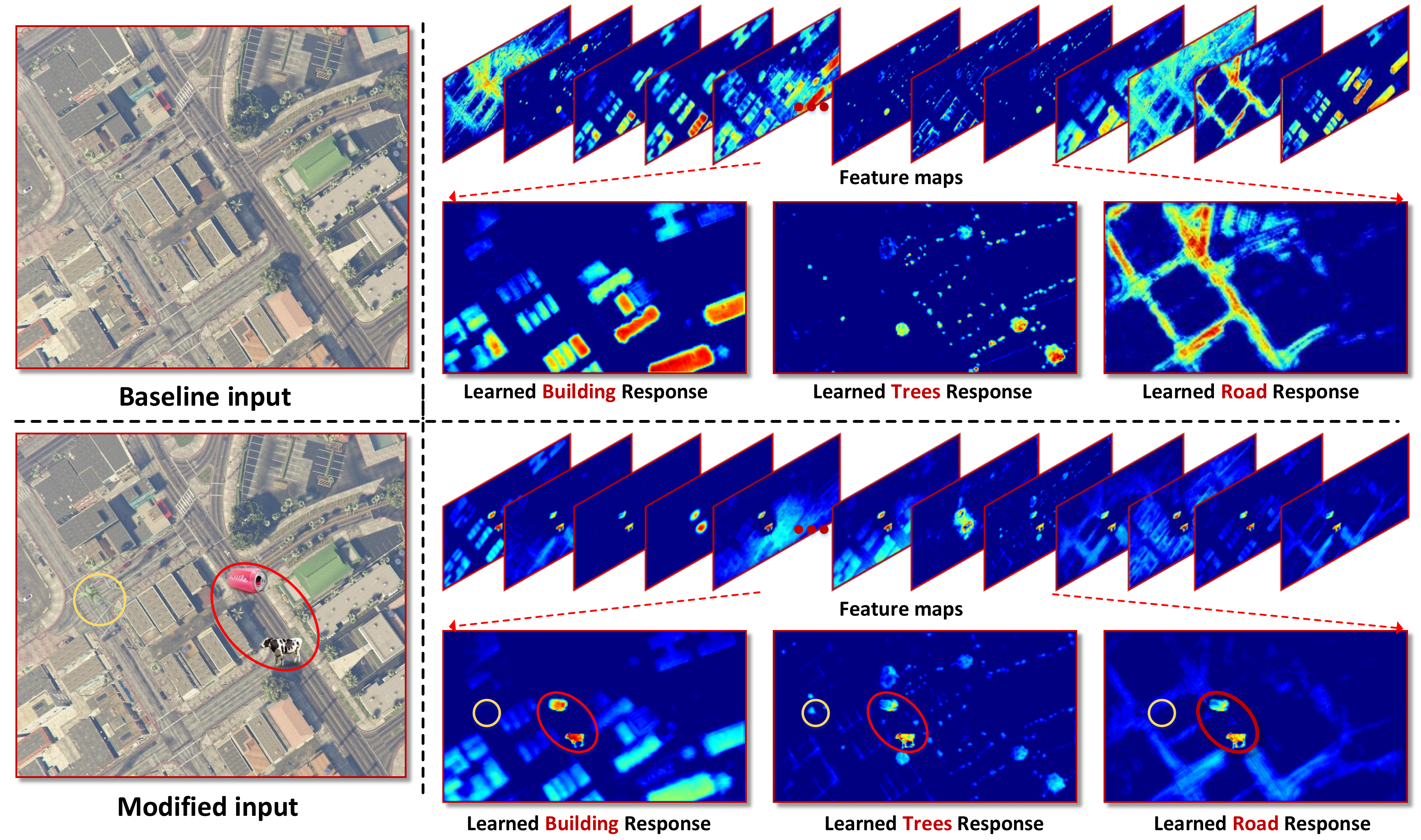}
	\caption{Out-of-distribution objects (\emph{cow} and \emph{bottle}) are highlighted in almost all the feature maps, which means these out-of-distribution objects are not recognized by the network. In contrast, the added \emph{tree} is only highlighted in the tree-selective feature maps. (Best viewed with zoom in)}
	\label{ood_response}
\end{figure*}
To quantify the selectivity of the network units, we compute both the class-selectivity \cite{MorcosBRB18} and the height-selectivity for each internal unit. In this work, we adapt the Swin Transformer \cite{liu2021swin} as the backbone for height estimation task. Denote each test sample in the dataset $D$ as $\mathbf{(x_i,y_i,h_i)}$, where ${\mathbf{x_i}} \in \R^{3,H,W}$ is the input image, ${\mathbf{y_i} \in \R^{H,W}}$, $\mathbf{h_i} \in \R^{H,W}$ are the ground truth semantic map and height map. $i \in \{1,...,N\}$. We upsample the $k_{th}$ feature map of the penultimate layer to the image size as $F_k(x_i) \in \R^{H,W}$. The class-selectivity and height-selectivity can be computed by:
\begin{equation}
\small
\label{CSHS}
\begin{gathered}
	\mathit{CR}^c_k=\frac{\sum_{1}^{N} S(F_k(\mathbf{x_i})\odot M^c_i)}{\sum_{1}^{N} S(M^c_i)};\quad \mathit{HR}^h_k=\frac{\sum_{1}^{N} S(F_k(\mathbf{x_i})\odot M^h_i)}{\sum_{1}^{N} S(M^h_i)},\\
	\mathit{CS}_k=\frac{|\mathit{CR}_k^{max}-\bar{\mathit{CR}}_k^{-max}|}{|\mathit{CR}_k^{max}+\bar{\mathit{CR}}_k^{-max}|};\quad \mathit{HS}_k=\frac{|\mathit{HR}_k^{max}-\bar{\mathit{HR}}_k^{-max}|}{|\mathit{HR}_k^{max}+\bar{\mathit{HR}}_k^{-max}|},
\end{gathered}
\end{equation}
where $\mathit{CR}^c_k$ and $\mathit{HR}^h_k$ are the average response of the units for different classes and height ranges. $h$ is the index of discretized height ranges, and $c$ is the index of semantic classes. $M^c$ is a binary mask indicating the pixels with semantic class $c$. $M^h$ is also a binary mask indicating the pixels in height range $h$. $S(\cdot)$ denotes the summation operation on all the elements of a matrix. We use $\odot$ to represent the element-wise multiplication.
With the defined average responses $\mathit{CR}^c_k$ and $\mathit{HR}^h_k$, class-selectivity and height-selectivity for the $k_{th}$ unit $\mathit{CS}_k$ and $\mathit{HS}_k$ can be computed. 
\begin{figure}
	\centering
	\includegraphics[width=0.48\textwidth]{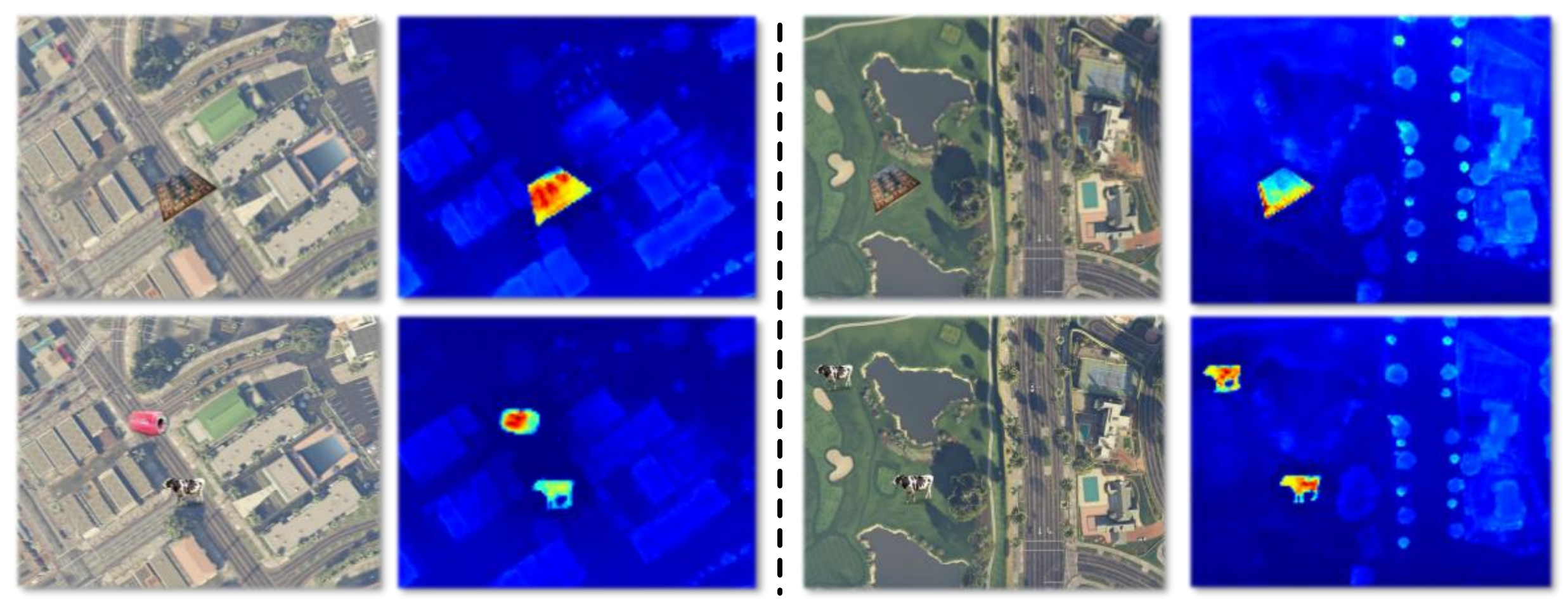}
	\caption{Visualization of out-of-distribution results. Out-of-distribution objects are highlighted in the anomaly response maps $R_{ood}$. (Best viewed in color).}
	\label{OOD_vis2}
\end{figure}

As defined in Eq. \ref{CSHS}, the value of $\mathit{CS}_k$ and $\mathit{HS}_k$ ranges from 0 to 1. Units of larger selectivity values tend to response preferably to specific semantic classes or height ranges. We have visualized the feature maps with high height-selectivity values in Fig. \ref{height-selectivity}, which shows high correlation between feature maps and different height ranges. Besides that, the quantified height-selectivity is also presented in Fig. \ref{HSB}. For both the real-world dataset and the synthetic dataset GTAH \cite{xiong2021benchmark}, there exist clear selectivity over different height ranges.

In Fig. \ref{CS-Real}, the class-selectivity of the Transformer-based MHE model is visualized. The class-selectivity of unit 255, 87 and 20 are larger than 0.8, and they are highly selective to categories \emph{Ground, High Vegetation,} and \emph{Building}, respectively. However, there are no clear selective units for \emph{Water} and \emph{Elevated Road}, since these categories have no clear differences compared with \emph{Ground} regarding height changes.

Based on the high class-selectivity in MHE networks, we propose a simple yet effective out-of-distribution method for MHE task. To better explain this method, we construct an out-of-distribution sample by adding some irrelevant objects, such as \emph{cow} and \emph{bottle} to the input image. As shown in Fig. \ref{ood_response}, we add a \emph{tree} (in-distribution) object as a comparison sample. Compared to the original input, we can see that the out-of-distribution objects (\emph{cow} and \emph{bottle}) are highlighted in almost all the feature maps. This means that these out-of-distribution objects are not recognized by the network. In contrast, the added \emph{tree} object is recognized by the MHE model, and is only highlighted in the tree-selective feature maps. 

To quantify this, we propose to compute the variance of pixels in the feature maps. Given the feature maps $F(\mathbf{x_i}) \in \R^{K,H,W}$, we first normalize it along the channels by $F'(\mathbf{x_i})=\frac{F(\mathbf{x_i})}{\sum_{1}^{K}F^k(\mathbf{x_i})}$. Then, we can compute the variance of $F'(\mathbf{x_i})$ for each pixel along the channel dimension. Finally, We can define the anomaly response map $R_{ood}=\bm{I}-var(F'(\mathbf{x_i}))$ as a measurement for out-of-distribution detection. Some visualization examples are shown in Fig. \ref{OOD_vis2}. It can be seen that out-of-distribution objects are detected with high response values.

\subsection{Instance-level Interpretation of MHE Models}
Considering that the learned units are highly selective to different semantic objects, it is natural to study the behavior of MHE networks by changing the semantics of the original input image. Thus, to keep the semantic concepts of original inputs, we choose to modify object instances of the original input image.

\begin{figure}
	\centering
	\includegraphics[width=0.5\textwidth]{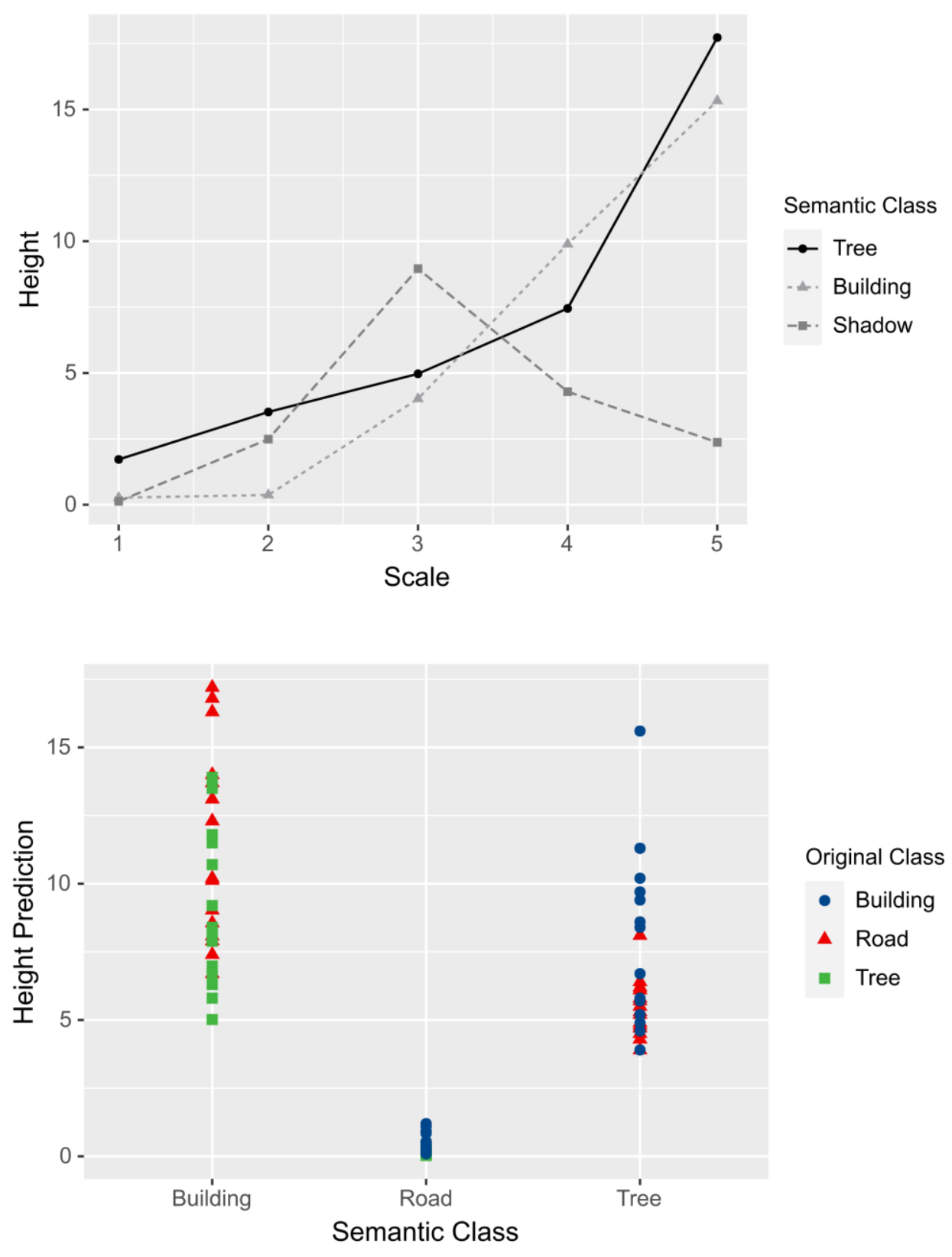}
	\caption{Visualization of the effects of object-level changes on the MHE networks.}
	\label{object_change}
\end{figure} 
Specifically, inspired by the cognition process of human beings, we explore the effects of changing the semantic class, scale and shadow state of object instances. Firstly, we define templates $t^c_s$ for each semantic category $c \in \{\emph{road},\emph{tree},\emph{building}\}$ and scale $s \in \{s_1,s_2,s_3,s_4,s_5\}$. Suppose the original scale (area) of the template for each category is $s_t$, then $\{s_1=0.3*st,s_2=1.0*st,s_3=1.5*st,s_4=2.5*st,s_5=3.0*st\}$.

To study the effects of changing semantic categories on height prediction, for each image $(\mathbf{x_i})$ in test dataset $D$, we select an appropriate patch with semantic class $\{\emph{road},\emph{tree},\emph{building}\}$ replaced with the predefined template $t^c_s$. The experimental results are shown in the left side of Fig. \ref{object_change}. We can see that the predicted heights change as expected after replacing objects of different semantic categories. This indicates that the height prediction results are highly correlated with the semantic categories of objects.

In order to explore the effects of changing object scales, for each image in the test set $D$, we first choose an appropriate location $\{p_x,p_y\}$ in the original image. Then, for each scale $s \in \{s_1,s_2,s_3,s_4,s_5\}$, we place the predefined template $t^c_s$ on the selected location. Since changing the scale of \emph{road} is not meaningful, we only conduct experiments on \emph{building} and \emph{tree} objects. The right side of Fig. \ref{object_change} presents the average height results for each scale. It can be seen that, in general, the larger the object scale is, the larger the height value is predicted. However, It is worth mentioning that, if the scale is too small, the network may not be able to recognize the object. Additionally, we also explore the effects of different scales of object shadows. From Fig. \ref{object_change} we can see that, the height predictions are not clearly correlated with the scale of shadows.

\subsection{Pixel-level Attribution Analysis of MHE Models}
Since MHE is a dense prediction task, interpreting the MHE networks for all pixels globally cannot provide meaningful interpretation information. 
From the unit-level and object-level analysis, we find that if the semantic class of an object cannot be recognized in the feature map, then the prediction result will be unreliable. This indicates that the existence of objects on feature maps is a fundamental and important factor for height prediction.
Thus, inspired by \cite{gu2021interpreting}, we design a local attribution analysis method based on feature existence and path integrated gradient \cite{sundararajan2017axiomatic} to interpret MHE networks in a pixel-level.

For an input image $\mathbf{x}$, $F_h{\mathbf{x}}$ represents the height prediction map predicted by MHE networks. 
We define $D_{px,py}(\mathbf{x})$ to quantify the existence of one local object as
\[D_{px,py}(\mathbf{x}) = \sum_{i\in [px,px+n],y\in [py,py+n]} F_h(\mathbf{x})_{ij}, \]
where $n$ is the window size of the selected patch. To mine the important pixels that account for the existence of one object, we use a whole black image as the baseline input $\mathbf{x'}$. In order to obtain the attribution map for the selected local patch, we need to integrate gradients along the gradually changing path from $\mathbf{x'}$ to $\mathbf{x}$. Note that, for the baseline image $\mathbf{x'}$, the output of the MHE network $F_h(\mathbf{x'})$ is not zero. 

Then the $i_{th}$ dimension of the integrated gradients for the local patch $\text{IG}$ can be computed by:
\begin{equation}
\label{EINT}
\text{IG}_i(x) := (\mathbf{x}-\mathbf{x'})\int_{\alpha=0}^{1} \frac{\partial D(F_h(\mathbf{x'}+\alpha (\mathbf{x}-\mathbf{x'})))}{\partial \mathbf{x}_i} \,d\alpha.
\end{equation}

In this work, we use the linear path function to define $\alpha$. To be specific, we interpolate $\alpha$ smoothly from 0 to 1 with $m$ steps, i.e., $\alpha=\frac{k}{m}$. In practice, we approximate the computation of the integral defined in Eq. \ref{EINT} by the following summation along $m$ steps:
\begin{equation}
	\text{IG}_i(x) := \frac{(\mathbf{x}-\mathbf{x'})}{m} \sum_{k=1}^{m} \frac{\partial D(F_h(\mathbf{x'}+\frac{k}{m}(\mathbf{x}-\mathbf{x'})))}{\partial \mathbf{x}_i} \,d\alpha.
\end{equation}

Empirically, setting $m$ to 100 works well for MHE networks in our experiments on both the real-world and synthetic datasets.

\section{Disentangled Latent Transformer}

Based on the finding of semantic selectivity, we pave a new perspective to understand the MHE model from multiple levels: 1) The semantic selectivity inspires the instance-level explanation and provides the reason why instance replacement makes sense; 2) The instance-level experiments further show that detecting the existence of local semantic objects by local attribution analysis is an effective way for pixel-level interpretation.

To make better use of the semantic selectivity, in this work, we further propose a disentangled latent Transformer (DLT) networks to explicitly model the representations for different semantic classes. By disentangling the deep neurons of MHE models into different semantic groups, a new unsupervised semantic segmentation method is motivated with the proposed DLT model. Specifically, DLT exploits a Transformer backbone to better learn the pairwise relations between different pixels. Next, a latent random variable is predicted for each semantic class, which can be used to generate semantic segmentation maps for a specific class.


We observe that different neurons of the final layer have specific response to different semantic classes, and there are more than one neurons responsible for a specific class. Thus, to enable the disentanglement representation learning, we first cluster these neurons into different semantic groups. Formally, given a pre-trained Transformer-based deep networks with $n$ neurons at the final layer, the weights $\{\bm{W_i}\in \R^{n_o\times k \times k}, i=1,...,n\}$ of the final layer are clustered into $K$ groups using K-means. Note that for each pre-trained deep models, we only need to cluster the neurons once and do not need to cluster the feature maps at each inference. Based on the predicted cluster indices, we can further obtain $K$ groups of feature maps $\{\bm{F_{ci}\in \R^{n_i,H,W}}, i=1,...,K\}$. Each semantic group contains $n_i$ feature maps, and $\sum_{i=1}^{K}n_i=n$.

Feature maps in the same semantic group are with similar semantic responses with slightly differences. To improve the reliability, we propose to treat the feature maps within the same semantic group as noisy observations, and model the true semantic response $\bm{F_v\in \R^{K\times H\times W}}$ as a latent variable remaining to be estimated. In practice, for each semantic class, we aim to sample the latent semantic responses from a posterior distribution $p(\bm{F_v}\mid x)$ to explicitly model the noise in a Bayesian probabilistic framework. 

As the true posterior is intractable to obtain, we adopt the variational inference \cite{kingma2013auto} to approximate $p(\bm{F_v}\mid x)$ with a simple distribution $q_{x}(\bm{F_v})$. Thus, the optimization goal is to minimize the Kullback-Leibler (KL) divergence $\mathcal{L}_{KL}=KL(q(\bm{F_v})\mid \mid p(\bm{F_v}|x))$. As pointed by \cite{kingma2013auto}, maximizing the likelihood distribution is equivalent to maximizing the variational evidence lower bound (ELBO), which can be defined as:
\begin{equation}
\begin{aligned}
\mathcal{L}_{KL} &=\sum_{\bm{F_v}} q(\bm{F_v} \mid x) \log \left(\frac{p(\bm{F_v}, x)}{q(\bm{F_v} \mid x)}\right) \\
&=\sum_{\bm{F_v}} q(\bm{F_v} \mid x) \log \left(\frac{p(x \mid \bm{F_v}) p(\bm{F_v})}{q(\bm{F_v} \mid x)}\right) \\
&=\sum_{\bm{F_v}} q_x(\bm{F_v}) \log \left(\frac{p(\bm{F_v})}{q_x(\bm{F_v})}\right)+\sum_{\bm{F_v}} q_x(\bm{F_v}) \log (p(x \mid \bm{F_v}))\\
&= \mathbb{E}_{q_{x}(\bm{F_v})}[\text{log}~p(\bm{F_c}|\bm{F_v})]-KL(q_{x}(\bm{F_v})||p(\bm{F_v})).
\end{aligned}
\end{equation}

In practice, the first expected log-likelihood term is used to minimize the feature reconstruction loss. Namely, we aim to reconstruct the original feature maps using the estimated latent variables $\bm{F_v}$. This can also be viewed as a feature distillation process. Then, the following KL divergence term is used to minimize the distance between $q_x(\bm{F_v})$ and the true posterior $p(\bm{F_v}\mid x)$. 


To train the whole model in an end-to-end manner using the gradient descent optimizer, we need to compute the gradients w.r.t. the parameters of distributions. For the sake of simplicity, we assume that the latent semantic responses $\bm{F_v}$ follow the isotropic Gaussian distribution with mean $\mu_i\in\R^{H,W}$ and variance ${\sigma_i}^2\in\R^{H,W}$.
However, the gradients of parameters $\{\mu_i,{\sigma_i}^2\}$ of the approximate distribution, which are predicted by deep networks, cannot be computed directly. Thus, we utilize the reparameterization trick to re-write $q_x(\bm{F_v})$ as ${\bm{F_{vi}}=\mu_i+{\sigma_i}\varepsilon, i=1,...,K}$, where $\varepsilon$ is the standard normal distribution, i.e., $\varepsilon \sim {\cal N}(\textbf{0},\textbf{I})$. With the reparameterization trick, the final loss function can be rewritten as:
\begin{equation}
\small
\begin{aligned}
{\cal L}(x^{(i)}) &= \frac{N}{M}\sum_{i=1}^{M}(\frac{1}{L}\sum_{l=1}^{L}\text{log}~p(y^{(i)}|x^{(i)},\mu)\\
&+ \sum\text{log}~p(\bm{F_c}^{(i)}|x^{(i)},\mu+{\sigma}\varepsilon)-KL(q_x(\bm{F_v}|x^{(i)})||p(\bm{F_v}))),
\end{aligned}
\label{ELBO}
\end{equation}
where $M$ is the size of mini-batch, and $N$ is the total number of samples. $L$ is the number of sampling process for Monte Carlo estimation. $y_i$ is the ground truth height maps for training the height estimation models.

For the prior distribution, we assume that the latent semantic responses follow isotropic Gaussian distribution with mean $\bm{\mu_i\in\R^{H,W}}$ and variance $\bm{{\sigma_i}^2\in\R^{H,W}}$. In this work, we use the mean of ${\bm{F_c}}$ as the prior mean of $\bm{F_v}$, and set the variance to be $\bm{I}$. Then the latent semantic responses $\bm{Fv_i}$ can be defined as 
\begin{equation}
\begin{aligned}
\bm{F_{vi}} \sim \mathcal{N}(\mu_pi,{\sigma_pi}^2),\\
\mu_pi = \frac{1}{ni}\sum {\bm{F_{ci}}}, \, \, \sigma_pi&=\bm{I},
\end{aligned}
\end{equation}

With the defined notations, the KL divergence loss in Eq. \ref{ELBO} for distribution approximation can be computed analytically as:
\begin{equation}
\small
KL(q_x(\bm{F_v})||p(\bm{F_v}))=-\frac{1}{2} \sum \left[1+\log \sigma_i^{2}-\sigma_{i}^{2}-(\mu_{i}-\mu_pi)^{2}\right].
\label{KL}
\end{equation}

After the training stage, the predicted mean of $\bm{F_v}$ can be used to generate the semantic segmentation maps for a specific class. In this work, we simply use OTSU \cite{otsu1979threshold} algorithm to binarize $\bm{F_v}$ into segmentation maps. Note that the final segmentation results are obtained from a MHE model without using any segmentation annotations.
\begin{figure}
	\centering
	\includegraphics[width=0.485\textwidth]{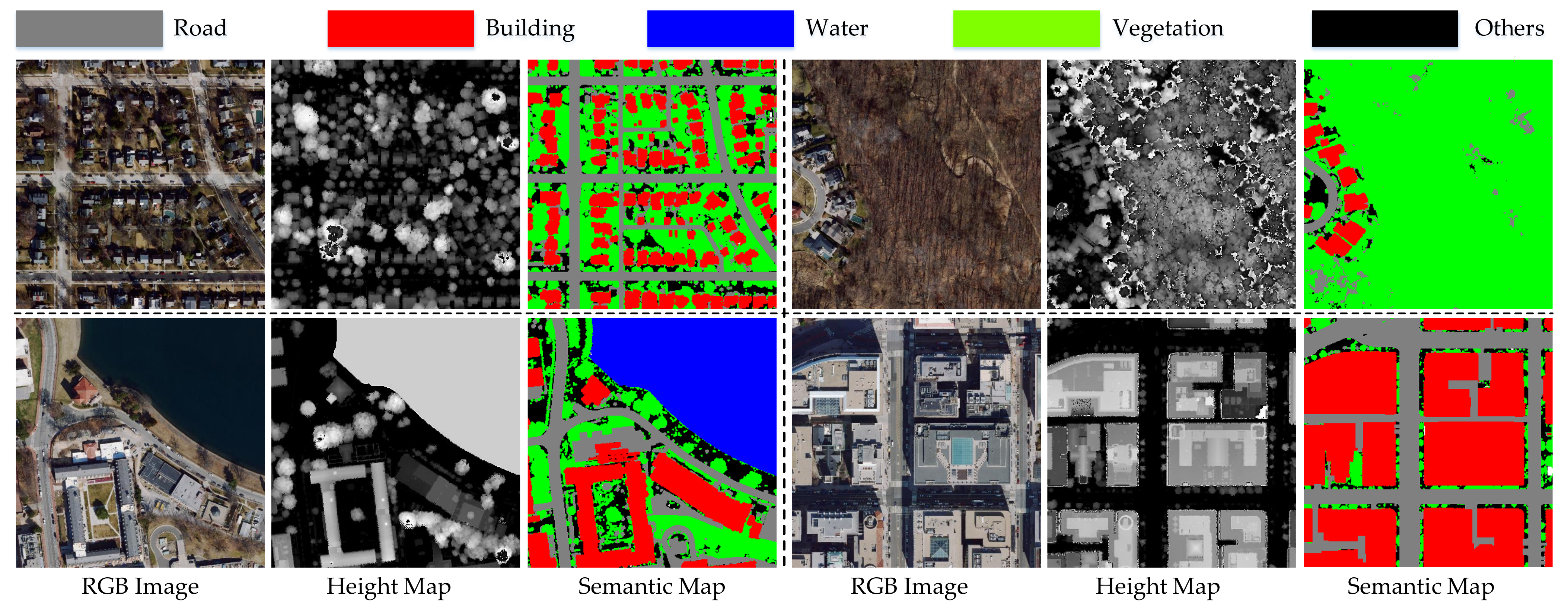}
	\caption{Visualization of the proposed Washington DC dataset. Patches of rural, forest, water, and urban are shown as examples. For each patch, the orthophoto, nDSM (height map), and semantic map are displayed.}
	\label{WDCData}
\end{figure} 
\section{Experiments}
\begin{figure*}[]
	\centering
	\includegraphics[width=0.968\textwidth]{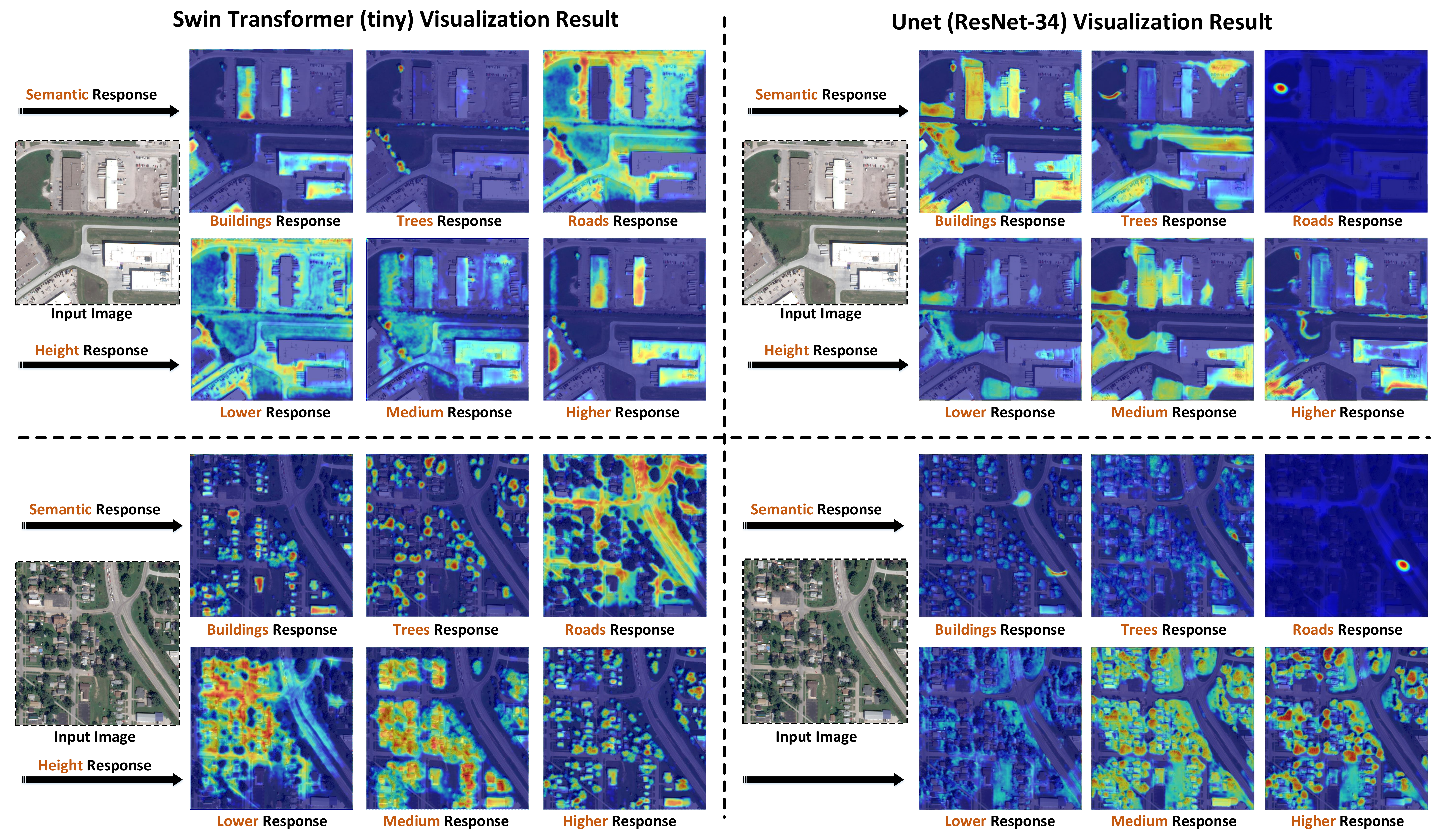}
	\caption{Transformer-based model can learn disentangled representations for different semantic objects. While the feature maps of UNet-based model are somehow entangled. (Best viewed in color and with zoom in)}
	\label{RMHEs}
\end{figure*}

In this section, we evaluate the proposed multi-level interpretation framework on different datasets. In addition, we use it to analyze the reason why Swin Transformer-based MHE model can outperform CNN-based MHE networks. 

\subsection{Datasets}
Three datasets are used in this work for the method evaluation. The first one is a real-world dataset from IEEE GRSS, Data Fusion Contest 2019 (DFC 2019) \cite{c6tm-vw12-19}. DFC 2019 contains 2,786 images with both the semantic segmentation label and normalized Digital Surface Model (nDSM). The second dataset is a synthetic dataset, called GTAH\footnote{\url{https://thebenchmarkh.github.io/}}, which is constructed using the game engine GTA V \cite{GTAV}. GTAH contains 85,881 images with the corresponding height map labels. Finally, we also evaluate the proposed method on our newly constructed WDC dataset in this work.

The proposed WDC dataset is processed based on open data at Open Data DC \footnote{\url{https://opendata.dc.gov/}}. The dataset covers whole Washington DC, USA. The aerial orthophoto was collected in 2021, and the nDSM was collected via LiDAR in 2020. There are 5 classes in the semantic map, namely ground, vegetation, buildings, water, and roads. Building footprints and roads in shape files are used to construct the semantic maps. For other classes, we refer to the labels in LiDAR point clouds. Some examples of the proposed dataset are visualized in Fig. \ref{WDCData}.

WDC dataset contains in total 2,159 patches of aerial orthophotos. Each image corresponds to a pixel-wise semantic map and a height map. The resolution of the image patch is 1024$\times$1024, and the GSD is 0.33 m. A variety of scenes are covered by the dataset, including urban, rural, forest, etc., thus, rich diversity of classes and heights.

\subsection{Implementation Details}
We implement the proposed method using Pytorch. On the GTAH dataset, we pre-train the Swin Transformer model with 100 epochs, and fine-tune the proposed DLT model for 20 epochs. For the U-Net model with ResNet-34 backbone, the code\footnote{\url{https://github.com/pubgeo/monocular-geocentric-pose}} from \cite{christie2020learning} is used. To optimize the Transformer-based models, AdamW \cite{loshchilov2018fixing} is used with an initial learning rate of 6e-5 for pre-training the deep models. For UNet-based models, Adam \cite{kingma2014adam} is used as the optimizer with an initial learning rate of 1e-4. For all the experiments, we set the batch size to 4.
More implementation details of this work are provided in the publicly available code.

We adapt the tiny version of Swin Transformer as the baseline method for MHE task. To compare with the CNN-based methods, we evaluate the UNet-based height estimation model with ResNet-34 and ResNet-101 backbones \cite{christie2020learning}.

\subsection{Results and Comparisons}
To evaluate the performance of the proposed methods on MHE, we follow the previous work \cite{xiong2021benchmark} and use four metrics for performance comparisons including Mean Absolute Error (MAE), Root Mean Squared Error (RMSE), Scale-Invariant RMSE (SI-RMSE) and Multi-scale Gradient Error (MSG). 
MAE is defined as $\text { MAE }=1 / n * \sum\left|y_{i}-\hat{y}_{i}\right|$, which is used to measure the mean absolute difference between the predicted values and the reference values. RMSE is defined as $\text { RMSE }=\sqrt{\Sigma\left(y_{i}-\hat{y}_{i}\right)^{2} / n}$, which is a commonly used measurement for regression task. To care more about the relative relations in the height maps, we also adopt the SI-RMSE and MSGE as metrics for model evaluation. SI-RMSE is defined as
\begin{equation}
\text{SI-RMSE} =\frac{1}{n} \sum_{i} R_{i}^{2}-\frac{1}{n^{2}}\left(\sum_{i} R_{i}\right)^{2}.
\end{equation}
The multi-scale gradient matching error, MSGE, can be formulated as:
\begin{equation}
\text{MSGE} =\frac{1}{M} \sum_{k=1}^{K} \sum_{i=1}^{M}\left(\left|\nabla_{x} R_{i}^{k}\right|+\left|\nabla_{y} R_{i}^{k}\right|\right).
\end{equation}

For the comparison of semantic segmentation, we use the commonly used Intersection over Union (IoU) and mean Intersection over Union (mIoU) as the evaluation metrics.
In the following sections, we first analyse and compare the performances of UNet and Swin Transformer models regarding the model interpretability. Both the unit-level and pixel-level interpretation methods are analysed. Then, we evaluate and compare the proposed DLT model on the performances of MHE and unsupervised semantic segmentation. Furthermore, we also compare the proposed new method with several state-of-the-art unsupervised segmentation methods.

\begin{table}[t]
	\centering
	\small
	\caption{Class-selectivity and response of different units of Swin Transformer on the DFC 2019 Dataset}
	\label{DFC1}
	\scalebox{0.88}{
	\begin{tabular}{c|c|c|c|c||c}
		\hline \hline
		Category    & Ground & High Vege. & Building & Water & Selectivity \\ \hline
		Unit 87  & 0.0833           & 0.7083               & 0.0485             & 0.1597          & \textbf{0.8893}      \\ \hline
		Unit 20  & 0.0433           & 0.2051               & 0.7241            & 0.0273          & \textbf{0.8359}      \\ \hline
		Unit 255 & 0.6724           & 0.1473               & 0.0606             & 0.1194          & \textbf{0.8302}      \\ \hline
		Unit 371 & 0.6412           & 0.1701               & 0.0443             & 0.1442          & {0.8089}      \\  \hline \hline
	\end{tabular}}
	\vspace{5px}
	\centering
	\caption{Class-selectivity and response of different units of UNet (ResNet-34) on the DFC 2019 Dataset}
	\label{DFC2}
	\scalebox{0.88}{
	\begin{tabular}{c|c|c|c|c||c}
		\hline \hline
		Category    & Ground & High Vege. & Building & Water & Selectivity \\ \hline
		Unit 17  & 0.7801           & 0.1826               & 0.0069             & 0.0302          & \textbf{0.8248}      \\ \hline
		Unit 29  & 0.7233           & 0.1873               & 0.0142             & 0.0751          & 0.8223      \\ \hline
		Unit 14 & 0.1419           & 0.4519               & 0.2801             & 0.1259          & 0.5635      \\ \hline
		Unit 16 & 0.1512           & 0.2738               & 0.3993             & 0.1756          & 0.5092      \\ \hline \hline
	\end{tabular}}
\end{table}

\subsubsection{Unit-level Analysis and Comparison}
In this section, we aim to analyze the reasons why the Transformer-based model can obtain better results than the UNet-based MHE method from the perspective of learned deep representations.
We compute the class-selectivity for two models and present the results in Table \ref{DFC1} and Table \ref{DFC2}. The last column of both tables indicates that Transformer-based MHE model has significantly stronger class-selectivity than UNet-based model. Besides, we also display the normalized average responses of these units. In general, the units of Transformer-based model are obviously more selective to different semantic categories. For a more clear comparison, the feature maps of two models are visualized in Fig. \ref{RMHEs}. We can see that the Swin transformer-based model can learn disentangled representations for different semantic objects: \emph{building, tree, road}. While the feature maps of UNet-based model are somehow entangled, i.e, buildings and trees are not well separated.

From the unit-level analysis, we can see the obvious superiority of the Transformer-based model in disentangled representation learning. When it comes to the quantitative performance, from the results in Table \ref{table:height_performance} we can see that the Swin Transformer-based MHE model can largely outperform the CNN-based UNet model. Especially for the GTAH dataset, Transformer-based model can reduce the SI-RMSE metric by almost 50\%.

\subsubsection{Pixel-level Analysis and Comparison}
For a more comprehensive comparison of the CNN and Transformer-based models, we conduct local attribution analysis towards a pixel-level understanding of them.
\begin{figure}
	\centering
	\includegraphics[width=0.465\textwidth]{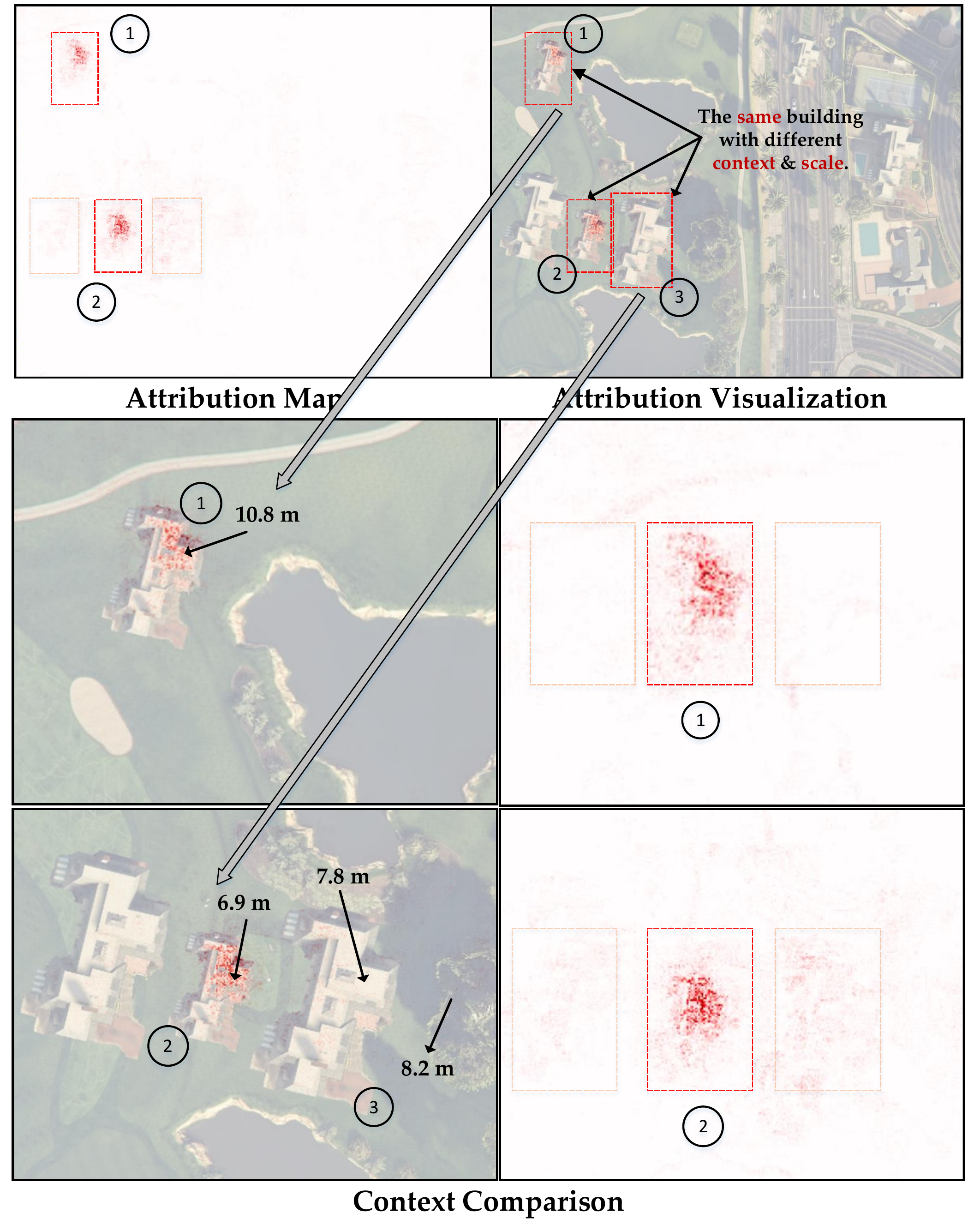}
	\caption{Comparison of the attribution maps for the same object template but different spatial context. (Best viewed in color and with zoom in)}
	\label{CSHOW}
\end{figure} 

\begin{figure}[t]
	\centering
	\includegraphics[width=0.49\textwidth]{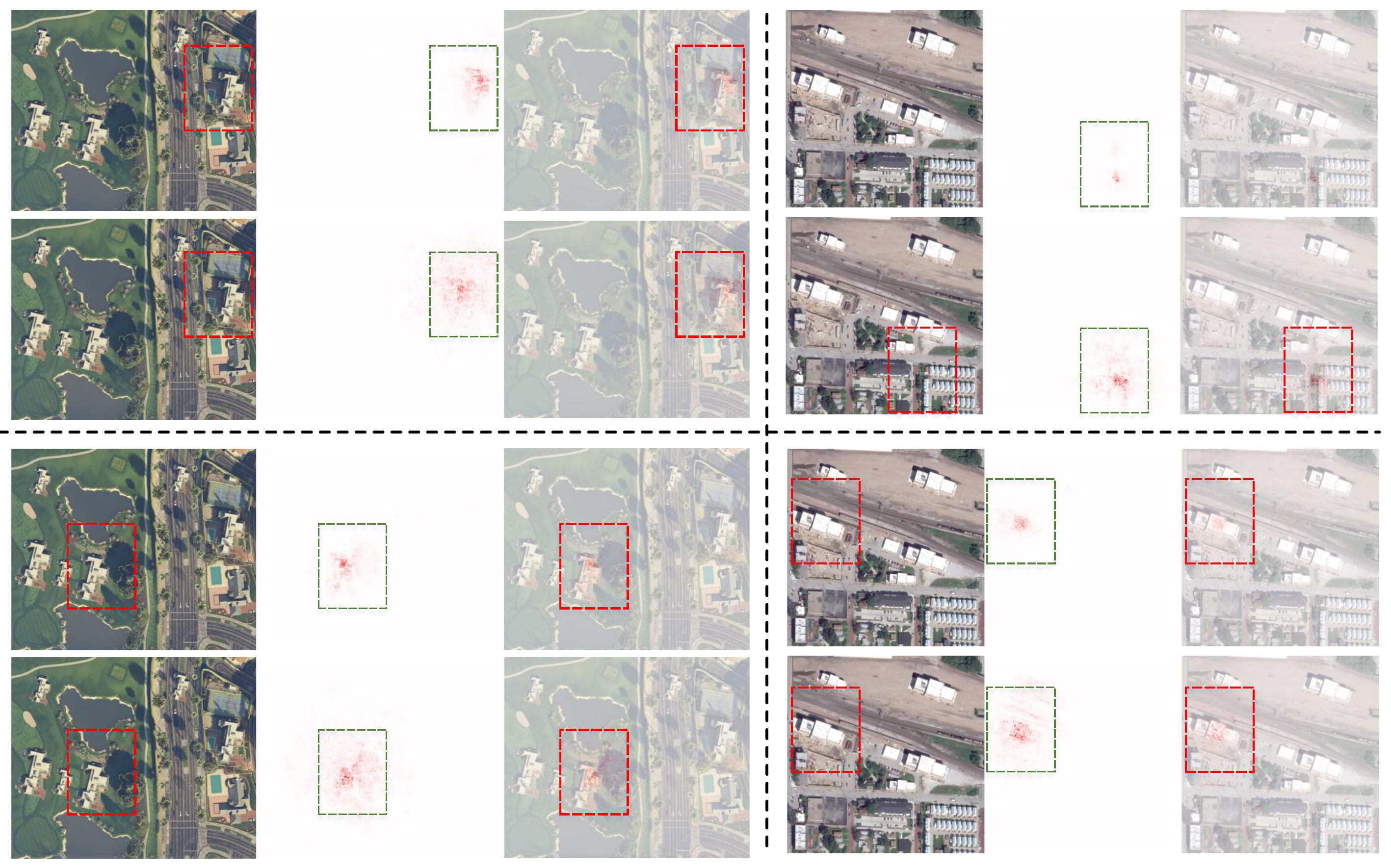}
	\caption{Comparison of the attribution maps for Transformer-based and UNet-based MHE networks. We can see that Transformer-based model can learn to exploit more effective and compact context. (Best viewed in color and with zoom in)}
	\label{AA_vis1}
\end{figure} 
To start with, we first analyze the importance of spatial context for height estimation task. For a clear presentation, we created an image containing four buildings with two different scales but the same template, as shown in Fig. \ref{CSHOW}. The first and second  buildings are of the same scale. However, their height predictions are quite different: 10.8 m vs. 6.9 m. While the ground truth is 7.2 m. This indicates that spatial context is also an important factor for MHE task in addition to the semantic class and object scale. From the attribution map, we can observe that the second building exploits much large context pixels on the surrounding buildings. Meanwhile, the height prediction for the third building is also influenced by the second building and the tree on the right.
\begin{table}[]
	\centering
	\small	
	\caption{Experimental results on the GTAH dataset in the zero-shot transfer setting. The best and second-best results are in \textcolor{blue}{blue} and \textcolor[rgb]{0,0.7,0.2}{green}.}
	\scalebox{0.87}{
		\begin{tabular}{c|c|c|c|c}
			\hline \hline
			\multirow{2}{*}{Methods} & \multicolumn{4}{c}{\textbf{Height Estimation}}                                                                                   \\ \cline{2-5} 
			& \textbf{MAE}    & \multicolumn{1}{l|}{\textbf{RMSE}} & \multicolumn{1}{l|}{\textbf{SI-RMSE}} & \multicolumn{1}{l}{\textbf{MSGE}} \\ \hline
			U-Net (ResNet34)         & 4.860          & 6.731                     & 39.511  & 3.357  \\ \hline
			U-Net (ResNet101)         & 3.216          & 5.807                     & 29.761  &2.968 \\ \hline
			Swin Transformer\cite{liu2021swin}   & \textcolor{blue}{2.970} & \textcolor{blue}{4.631}            & \textcolor{blue}{21.161}                               & \textcolor{blue}{2.403}                             \\ \hline
			Swin (16 Units)      & 3.093          & 4.750                             & {22.039}                      & {2.527}                    \\ \hline
			Swin (16 Units)+Distillation       & 3.194          & 4.778                             & 22.313                               & 2.571                             \\ \hline
			Swin (16 Units)+DLT(Ours)   & \textcolor[rgb]{0,0.7,0.2}{2.974}          & \textcolor[rgb]{0,0.7,0.2}{4.731}                             & \textcolor[rgb]{0,0.7,0.2}{21.614}                               & \textcolor[rgb]{0,0.7,0.2}{2.516}                             \\ 
		 \hline \hline
		\end{tabular}
	}
\label{table:height_performance}
\end{table}

Since context information is an important factor for MHE, we visualize the attribution maps of Transformer-based and UNet-based models to compare their spatial contexts. As shown in Fig. \ref{AA_vis1}, the upper rows show that the Transformer-based model tends to use the pixels of the specific object. Thus, the context is more compact than the UNet-based model. In contrast, the UNet-based MHE model usually contains many irrelevant pixels, which makes it not robust in complex scenes. From this perspective, we can conclude that the self-attention mechanism in Transformer models is helpful for learning disentangled representation and exploiting more effective spatial context.

\begin{table}[]
	\centering
	\caption{Height estimation results of different methods on the DFC 2019 dataset. The best and second-best results are in \textcolor{blue}{blue} and \textcolor[rgb]{0,0.7,0.2}{green}.}
	\scalebox{0.82}{
	\begin{tabular}{c|cccc}
\hline \hline
\multirow{2}{*}{Method}       & \multicolumn{4}{c}{Height Estimation}                                                                             \\ \cline{2-5} 
                              & \multicolumn{1}{c|}{MAE}   & \multicolumn{1}{l|}{RMSE}  & \multicolumn{1}{l|}{SI-RMSE} & \multicolumn{1}{l}{MSGE} \\ \hline
U-Net (ResNet-34)             & \multicolumn{1}{c|}{1.432} & \multicolumn{1}{c|}{2.417} & \multicolumn{1}{c|}{7.114}   & 4.015                    \\ \hline
U-Net (ResNet-101)            & \multicolumn{1}{c|}{1.323} & \multicolumn{1}{c|}{2.314} & \multicolumn{1}{c|}{6.415}   & 3.393                    \\ \hline
Swin Transformer              & \multicolumn{1}{c|}{\textcolor{blue}{1.236}} & \multicolumn{1}{c|}{\textcolor{blue}{2.116}} & \multicolumn{1}{c|}{\textcolor{blue}{5.326}}   & 3.031                    \\ \hline
SwinT (16 Units)              & \multicolumn{1}{c|}{1.288} & \multicolumn{1}{c|}{2.206} & \multicolumn{1}{c|}{5.674}   & \textcolor{blue}{2.975}                    \\ \hline
SwinT (16 Units)+Distillation & \multicolumn{1}{c|}{1.414} & \multicolumn{1}{c|}{2.240} & \multicolumn{1}{c|}{5.947}   & \textcolor[rgb]{0,0.7,0.2}{2.959}                    \\ \hline
SwinT (16 Units)+DLT (Ours)   & \multicolumn{1}{c|}{\textcolor[rgb]{0,0.7,0.2}{1.274}} & \multicolumn{1}{c|}{\textcolor[rgb]{0,0.7,0.2}{2.179}} & \multicolumn{1}{c|}{\textcolor[rgb]{0,0.7,0.2}{5.589}}   & 3.016                    \\ \hline \hline
\end{tabular}}
\label{table::DFC_comparison}
\end{table}

\begin{table}[]
	\centering
	\small	
	\caption{Unsupervised semantic segmentation results on the DFC 2019 dataset. The best results are presented in bold.}
	\scalebox{0.82}{
		\begin{tabular}{c|ccc}
\hline \hline
\multirow{2}{*}{Method}       & \multicolumn{3}{c}{Semantic Segmentation}                                      \\ \cline{2-4} 
                              & \multicolumn{1}{c|}{Building(IoU)} & \multicolumn{1}{c|}{Tree(IoU)} & Mean IoU \\ \hline
PiCIE                         & \multicolumn{1}{c|}{0.150}         & \multicolumn{1}{c|}{0.280}     & 0.215    \\ \hline
IIC                           & \multicolumn{1}{c|}{0.157}         & \multicolumn{1}{c|}{0.275}     & 0.216         \\ \hline
MaskContrast                  & \multicolumn{1}{c|}{0}             & \multicolumn{1}{c|}{0.0001}         & 0.0001    \\ \hline \hline
U-Net (16 Units)              & \multicolumn{1}{c|}{0.092}         & \multicolumn{1}{c|}{0.077}     & 0.0845   \\ \hline
SwinT (16 Units)              & \multicolumn{1}{c|}{0.269}              & \multicolumn{1}{c|}{0.348}          & 0.309         \\ \hline
SwinT (16 Units)+Distillation & \multicolumn{1}{c|}{0.281}              & \multicolumn{1}{c|}{0.357}          & 0.319         \\ \hline
SwinT (16 Units)+DLT (Ours)   & \multicolumn{1}{c|}{\textbf{0.306}}              & \multicolumn{1}{c|}{\textbf{0.389}}          &  \textbf{0.348}        \\ \hline \hline
\end{tabular}}
\label{table:DFC_seg_performance}
\end{table}

\subsection{Disentangled Latent Transformer}
In the following sections, we evaluate and compare the performances of our proposed DLT model with previous state-of-the-art (SOTA) methods on both the height estimation and unsupervised semantic segmentation tasks.
\subsubsection{Height Estimation Performance Comparison}
\begin{table}[]
	\centering
	\caption{Height estimation results of different methods on the proposed WDC 2019 dataset. The best and second-best results are in \textcolor{blue}{blue} and \textcolor[rgb]{0,0.7,0.2}{green}.}
	\scalebox{0.84}{
	\begin{tabular}{c|cccc}
\hline \hline
\multirow{2}{*}{Method}       & \multicolumn{4}{c}{Height Estimation}                                                                             \\ \cline{2-5} 
                              & \multicolumn{1}{c|}{MAE}   & \multicolumn{1}{l|}{RMSE}  & \multicolumn{1}{l|}{SI-RMSE} & \multicolumn{1}{l}{MSGE} \\ \hline
U-Net (ResNet-34)             & \multicolumn{1}{c|}{4.860} & \multicolumn{1}{c|}{6.271} & \multicolumn{1}{c|}{39.670}   & 6.924                    \\ \hline
U-Net (ResNet-101)            & \multicolumn{1}{c|}{4.483} & \multicolumn{1}{c|}{5.832} & \multicolumn{1}{c|}{35.048}   & 6.245                    \\ \hline
Swin Transformer              & \multicolumn{1}{c|}{\textcolor{blue}{4.146}} & \multicolumn{1}{c|}{\textcolor{blue}{5.403}} & \multicolumn{1}{c|}{\textcolor{blue}{31.960}}   & \textcolor{blue}{5.857}                    \\ \hline
SwinT (16 Units)              & \multicolumn{1}{c|}{4.412} & \multicolumn{1}{c|}{5.734} & \multicolumn{1}{c|}{35.608}   & 6.136                    \\ \hline
SwinT (16 Units)+Distillation & \multicolumn{1}{c|}{4.462} & \multicolumn{1}{c|}{5.753} & \multicolumn{1}{c|}{35.948}   & 6.357                    \\ \hline
SwinT (16 Units)+DLT (Ours)   & \multicolumn{1}{c|}{\textcolor[rgb]{0,0.7,0.2}{4.346}} & \multicolumn{1}{c|}{\textcolor[rgb]{0,0.7,0.2}{5.709}} & \multicolumn{1}{c|}{\textcolor[rgb]{0,0.7,0.2}{33.89}}   & \textcolor[rgb]{0,0.7,0.2}{6.094}             \\ \hline \hline
\end{tabular}}
\label{table::WDC_comparison}
\end{table}

\begin{table}[]
	\centering
	\small	
	\caption{Unsupervised semantic segmentation results on the proposed WDC dataset. The best results are presented in bold.}
	\scalebox{0.82}{
		\begin{tabular}{c|ccc}
\hline \hline
\multirow{2}{*}{Method}       & \multicolumn{3}{c}{Semantic Segmentation}                                      \\ \cline{2-4} 
                              & \multicolumn{1}{c|}{Building(IoU)} & \multicolumn{1}{c|}{Tree(IoU)} & Mean IoU \\ \hline
PiCIE                         & \multicolumn{1}{c|}{0.278}         & \multicolumn{1}{c|}{0.221}     & 0.250    \\ \hline
IIC                           & \multicolumn{1}{c|}{0.201}         & \multicolumn{1}{c|}{0.289}     & 0.245    \\ \hline
MaskContrast                  & \multicolumn{1}{c|}{0.002}         & \multicolumn{1}{c|}{0.368}     & 0.185    \\ \hline \hline
U-Net (16 Units)              & \multicolumn{1}{c|}{0.113}         & \multicolumn{1}{c|}{0.127}     & 0.120   \\ \hline
SwinT (16 Units)              & \multicolumn{1}{c|}{0.336}              & \multicolumn{1}{c|}{0.389}          & 0.363          \\ \hline
SwinT (16 Units)+Distillation & \multicolumn{1}{c|}{0.348}              & \multicolumn{1}{c|}{0.397}          & 0.373         \\ \hline
SwinT (16 Units)+DLT (Ours)   & \multicolumn{1}{c|}{\textbf{0.365}}              & \multicolumn{1}{c|}{\textbf{0.412}}          & \textbf{0.389}         \\ \hline \hline
\end{tabular}}
\label{table:WDC_seg_performance}
\end{table}

\textbf{Comparisons on the GTAH Dataset.} 
As we find that more than one neurons are responsible for the same semantic class, we propose to reduce the number of neurons in the final layer by removing redundant neurons.
Thus, we can make each neuron in the last layer to be only sensitive to a specific semantic class or height range. With this neuron compression process, we can obtain a more compact deep neural network with higher interpretability. As presented in Table  \ref{table::DFC_comparison}, "SwinT (16 Units)" denotes the model where the units (number of channels) of the last layer are reduced from 512 to 16.
\begin{figure*}[]
	\centering
	\includegraphics[width=0.982\textwidth]{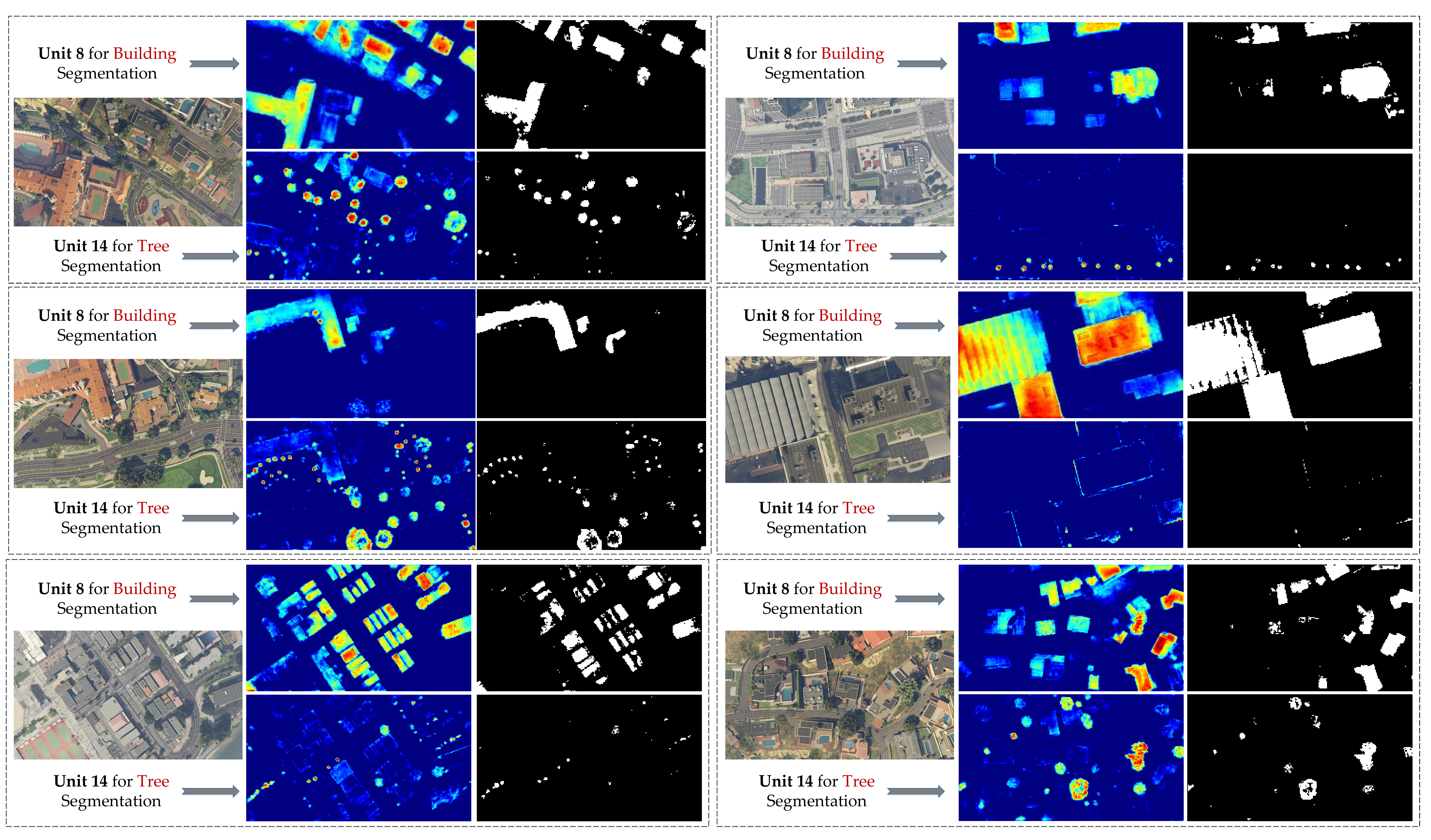}
	\caption{Unsupervised semantic segmentation results of the proposed ``SwinT (16 Units)+DLT'' on the GTAH dataset. The feature maps and segmentation maps of Unit 8 (building) and Unit 14 (tree) are shown in this figure. (Best viewed with zoom in)}
	\label{GTAH_vis}
\end{figure*}

However, reducing the number of neurons from 512 to 16 significantly limits the capacity of the model, which also affects the final height estimation performance of the model. Therefore, the performance of model ``SwinT (16 units)'' is lower than that of the original model (512 units) on the GTAH dataset. To cope with this issue, we propose to distill the feature maps of the  original model to the compressed one by encouraging the high feature similarity between them. However, it can be seen from the experimental results that the model "SwinT (16 Units)+Distillation" with a simple feature distillation does not lead to performance improvement. The reason is that simple feature distillation limits the expressiveness of the model due to the noise in the features. The DLT model proposed in this paper, on the other hand, better captures the noise in the features by explicitly modeling the features as latent random variable, thus obtaining very similar results to the original model. The experimental results in the table demonstrate the effectiveness of the proposed DLT model. We also visualize some unsupervised segmentation results in Fig. \ref{GTAH_vis}.
\begin{figure*}[]
	\centering
	\includegraphics[width=1.0\textwidth]{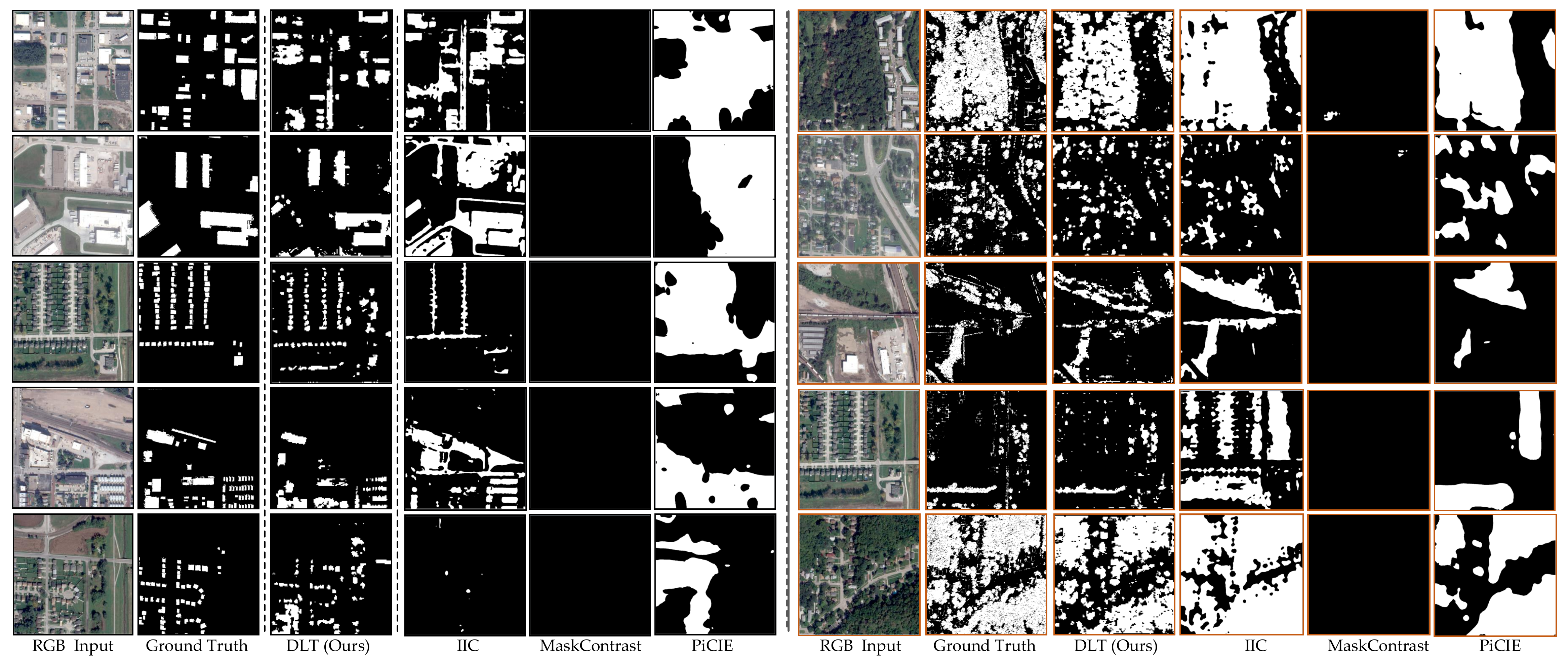}
	\caption{Unsupervised semantic segmentation results of the proposed ``SwinT (16 Units)+DLT'' on the DFC 2019 dataset. The segmentation maps of Unit 3 (building, the left side) and Unit 7 (tree, the right side) are shown in this figure. Compared with its counterparts, our DLT can obtain clearly better segmentation results. (Best viewed with zoom in)}
	\label{uss_vis}
\end{figure*}

\textbf{Comparisons on the DFC 2019 Dataset.} Similar to the GTAH dataset, we also compare the proposed method with other SOTA methods on the DFC 2019 dataset. From the results in Table \ref{table::DFC_comparison}, it can be clearly seen that our proposed "SwinT (16 Units)+DLT" with only 16 units can achieve competitive height estimation performance compared with the original Transformer model. This indicates the effectiveness of the proposed DLT model on the real-world dataset. From the results, we can also observe that Transformer-based models can obtain better performance when compared with CNN-based UNet models. This comparison demonstrates that Transformer-based deep models can be more suitable than CNN-based models for the pixel-wise dense regression task.

\textbf{Comparisons on the WDC Dataset.} 
On the proposed WDC dataset, we can also observe that Transformer-based models have advantages in terms of height estimation performance. The proposed ``SwinT (16 Units)+DLT" obtains similar performance compared with the ``Swin Transformer'' model. It shows that the compressed DLT model can improve the interpretability and reduce the computational complexity of the original Transformer model, with no loss of accuracy.

\subsubsection{Unsupervised Segmentation Results Comparison}
We also compare our results with three unsupervised semantic segmentation networks, including MaskContrast \cite{unsup2021mc}, PiCIE \cite{unsup2021picie}, and IIC \cite{unsup2019iic}. MaskContrast uses saliency as the prior knowledge for contrastive learning, and gains SOTA performance for natural images. However, remote sensing images are quite different from natural images in terms of contexts and scales. In our experiments, we find that the saliency-based method fails to distinguish foreground and background on DFC dataset. Thus, Almost all the image patches are assigned with only the ground label. PiCIE considers the photometric and geometric invariance of pixel embeddings. It also suffers from the difficulties caused by the differences between natural and remote sensing images. Though IoU socres are high, the predictions do not capture the layouts of input images. Of these three methods, The mutual information maximization-based method, IIC, is the only one that has proven to be effective on remote sensing images. From the visualization, it predicts well in terms of the layout. 

For the unsupervised segmentation task, we only compare the performance on \textbf{Building} and \textbf{Tree} classes. The reason is that buildings and trees are objects that are important for urban planning, climate change and disaster monitoring. Traditional supervised learning-based methods usually require a large amount annotation data for model training. However, annotating pixel-level semantic labels is expensive for large-scale or even global-scale Earth observation applications. 

\begin{figure}[t]
	\centering
	\includegraphics[width=0.50\textwidth]{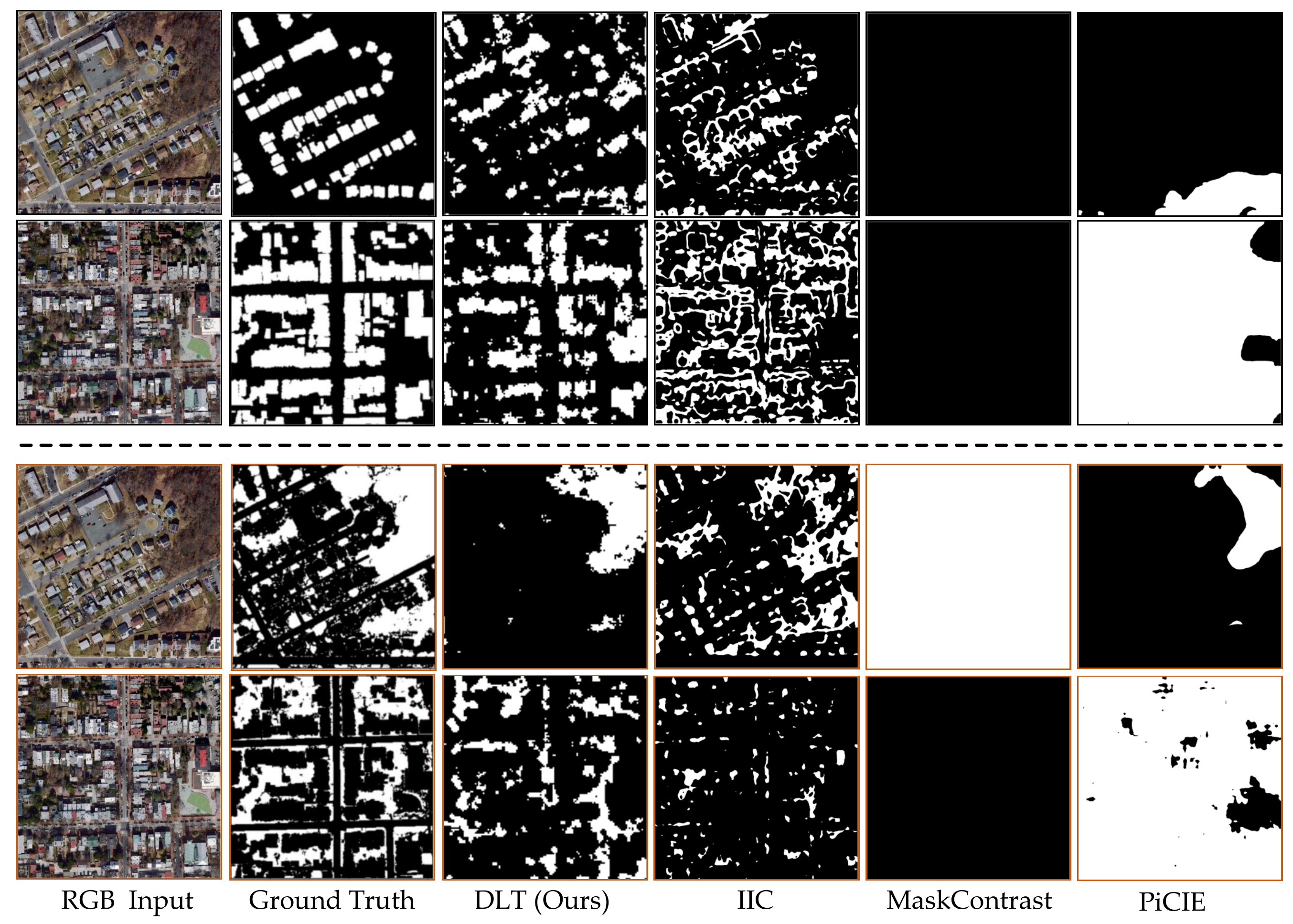}
	\caption{Unsupervised semantic segmentation results of the proposed ``SwinT (16 Units)+DLT'' on the WDC dataset. The segmentation maps of Unit 12 (building, upper two rows) and Unit 0 (tree, lower two rows) are shown in this figure. Compared with its counterparts, our DLT can obtain clearly better segmentation results. (Best viewed with zoom in)}
	\label{WDC_vis}
\end{figure}

In contrast, our proposed DLT model can obtain the pixel-level semantic segmentation maps using only the height maps as supervision, which has great potential for large-scale applications. The results in Table \ref{table:DFC_seg_performance} show that DLT can outperform existing unsupervised semantic segmentation methods clearly by a large margin. For example, compared with IIC, DLT can increase the IoU metric from 0.157 to 0.306. As shown in Table \ref{table:WDC_seg_performance}, on the WDC Dataset, we can also observe that compared with MaskContrast \cite{unsup2021mc}, PiCIE \cite{unsup2021picie}, and IIC \cite{unsup2019iic}, the proposed DLT model can achieve much better segmentation performance. The aforementioned experimental results have demonstrated the effectiveness of our proposed DLT model on both height estimation and unsupervised segmentation tasks.
In Fig.\ref{uss_vis} and Fig. \ref{WDC_vis}, we visualize some qualitative examples of our DLT method and its counterparts for a clear comparison.

\section{Conclusion}
In this paper, we explore how deep networks predict height from single images. To understand the MHE networks comprehensively, three levels of interpretation are studied: 1) Neurons: unit-level network dissection; 2) Instances: object-level interpretation; 3) Attribution: pixel-level analysis. From our experiments, we find that: 1) Deep neurons learned by MHE networks have high selectivity to both the height ranges and the semantic categories; 2) Based on the class-selectivity, we propose a simple yet effective out-of-distribution detection method for MHE; 3) Semantic class, object scale and spatial context are the main factors influencing the prediction results; 4) Transformer based networks have stronger height and class selectivity and can use spatial context more effectively than CNNs. Based on the understandings of MHE models, a disentangled latent Transformer model is designed towards a more compact and explainable deep networks for monocular height estimation. The proposed DLT model has shown superiority on the unsupervised semantic segmentation task. Furthermore, we also release a new dataset for both semantic segmentation and height estimation. This work provides novel insights for future works to better understand and design MHE models.
		

\clearpage

\bibliographystyle{IEEEtran}
\bibliography{IEEEabrv,refs}

\begin{IEEEbiography}[{\includegraphics[width=1in,height=1.25in,clip,keepaspectratio]{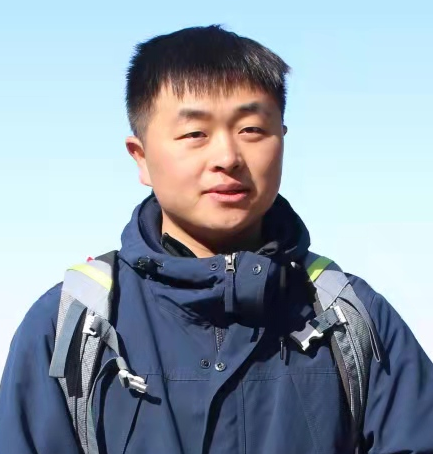}}]{Zhitong Xiong} received the Ph.D. degree in computer science and technology from the Northwestern Polytechnical University, Xi’an, China, in 2021. He is currently a Postdoc with the Data Science in Earth Observation, Technical University of Munich (TUM), Munich, Germany. His research interests include computer vision, machine learning and remote sensing.
\end{IEEEbiography}
\begin{IEEEbiography}[{\includegraphics[width=1in,height=1.25in,clip,keepaspectratio]{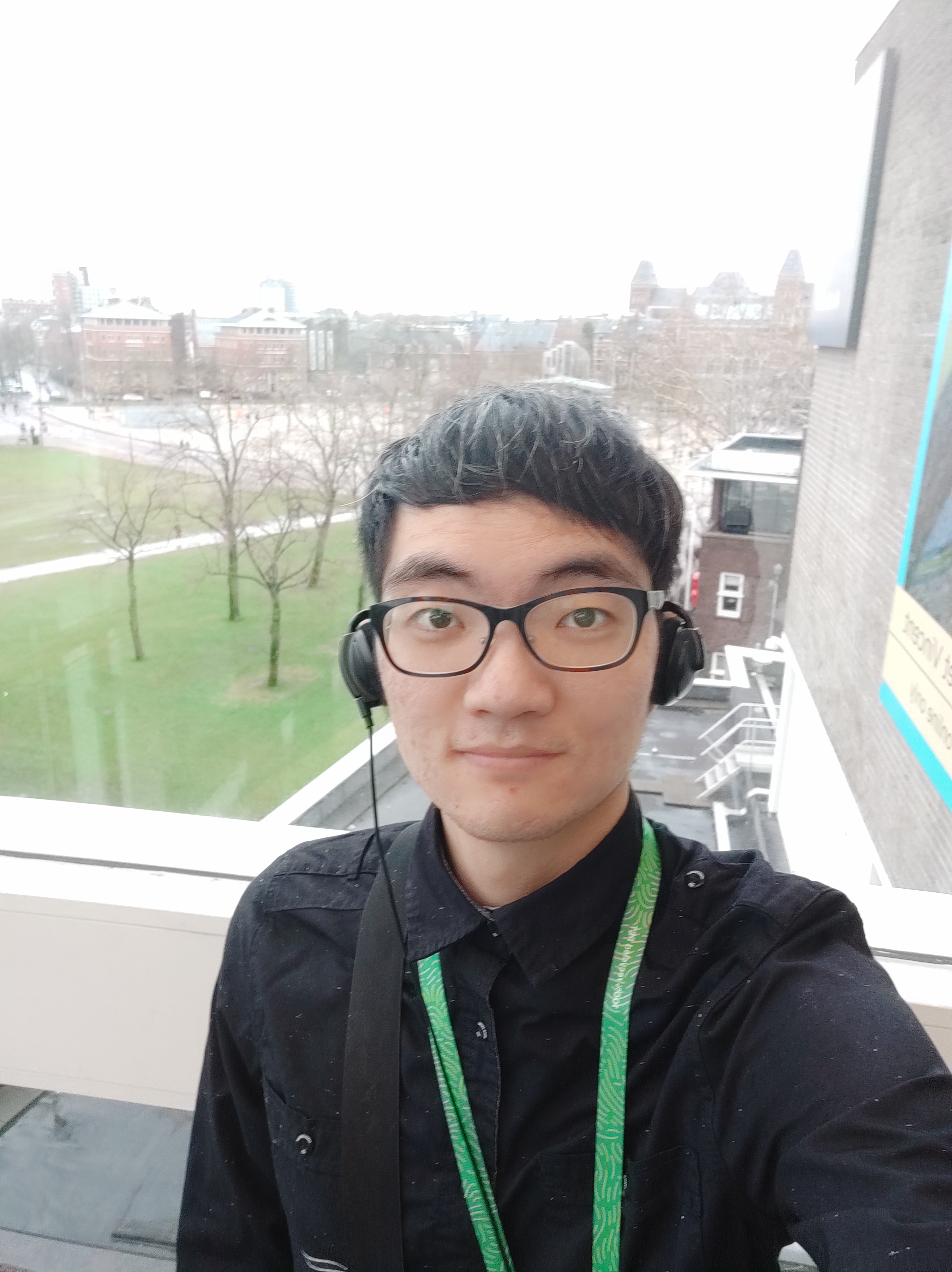}}]{Sining Chen} received the master's degree in Earth-oriented Space Science and Technology (ESPACE) from Technical University of Munich (TUM), Germany. He is currently a Ph.D. student with the Remote Sensing Technology Institute, German Aerospace Center (DLR) and Data Science in Earth Observation, Technical University of Munich (TUM), Germany. His research interests include deep learning and building reconstruction.
\end{IEEEbiography}

\begin{IEEEbiography}[{\includegraphics[width=1in,height=1.25in,clip,keepaspectratio]{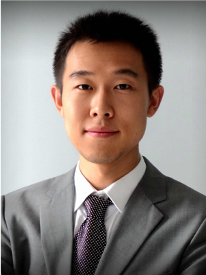}}]{Yilei Shi}(M'18) received his Diploma (Dipl.-Ing.) degree in Mechanical Engineering, his Doctorate (Dr.-Ing.) degree in Engineering from Technical University of Munich (TUM), Germany. In April and May 2019, he was a guest scientist with the department of applied mathematics and theoretical physics, University of Cambridge, United Kingdom. He is currently a senior scientist with the Chair of Remote Sensing Technology, Technical University of Munich.
\end{IEEEbiography}

\begin{IEEEbiography}[{\includegraphics[width=1in,height=1.25in,clip,keepaspectratio]{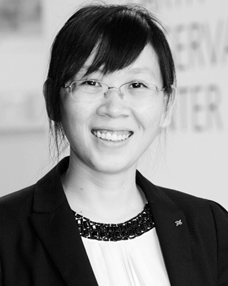}}]{Xiaoxiang Zhu}.
(S’10-M’12-SM’14-F’20) received the M.Sc., Dr.Ing., and Habilitation degrees in signal
processing from the Technical University of Munich (TUM), Munich, Germany, in 2008, 2011, and 2013,
respectively. She is currently the Professor for Data Science in Earth Observation at Technical University of Munich (TUM) and the Head of the Department ``EO Data Science'' at the Remote Sensing Technology Institute, German Aerospace Center (DLR). Since 2019, Zhu is a co-coordinator of the Munich Data Science Research School (www.mu-ds.de). Since 2019 She also heads the Helmholtz Artificial Intelligence -- Research Field ``Aeronautics, Space and Transport". Since May 2020, she is the director of the international future AI lab "AI4EO -- Artificial Intelligence for Earth Observation: Reasoning, Uncertainties, Ethics and Beyond", Munich, Germany. Since October 2020, she also serves as a co-director of the Munich Data Science Institute (MDSI), TUM. Prof. Zhu was a guest scientist or visiting professor at the Italian National Research Council (CNR-IREA), Naples, Italy, Fudan University, Shanghai, China, the University  of Tokyo, Tokyo, Japan and University of California, Los Angeles, United States in 2009, 2014, 2015 and 2016, respectively. She is currently a visiting AI professor at ESA's Phi-lab. Her main research interests are remote sensing and Earth observation, signal processing, machine learning and data science, with a special application focus on global urban mapping. 
\end{IEEEbiography}

\ifCLASSOPTIONcaptionsoff
  \newpage
\fi

\end{document}